%% file: main.tex
\documentclass[10pt,twocolumn,letterpaper]{article}

\usepackage[pagenumbers]{cvpr} 

\input{preamble}

\definecolor{cvprblue}{rgb}{0.21,0.49,0.74}
\usepackage[pagebackref,breaklinks,colorlinks,citecolor=cvprblue]{hyperref}
\usepackage{bbding} 
\usepackage{makecell}
\usepackage{CJK}

\usepackage{graphicx}
\usepackage{caption}
\usepackage{subcaption}
\usepackage{multirow}
\usepackage{algorithm}
\usepackage{algorithmic}
\usepackage{mathrsfs}

\def\paperID{3642} 
\def\confName{CVPR}
\def\confYear{2024}

\title{Instance by Instance: An Iterative Framework for Multi-instance 3D Registration}

\author{Xinyue Cao\textsuperscript{1}, Xiyu Zhang\textsuperscript{1}, Yuxin Cheng\textsuperscript{2}, Zhaoshuai Qi\textsuperscript{1}, Yanning Zhang\textsuperscript{1}, Jiaqi Yang\textsuperscript{1,*}\\
$^{1}$School of Computer Science, Northwestern Polytechnical University, China\\
$^{2}$Department of Electronic and Computer Engineering, Chinese University of Hong Kong, China\\
{\tt\small $^{1}\{$caoxinyue,2426988253$\}$@mail.nwpu.edu.cn; $\{$zhaoshuaiqi1206,ynzhang,jqyang$\}$@nwpu.edu.cn;}\\
{\tt\small $^{2}$yuxin.cheng@link.cuhk.edu.cn}}

\begin{document}
\maketitle
\input{sec/0_abstract}    
\input{sec/1_introduction}
\input{sec/2_related_work}
\input{sec/3_IBI_framework}
\input{sec/4_IBI-S2DC}
\input{sec/5_experiment}
\input{sec/6_conclusion}

{
    \small
    \bibliographystyle{ieeenat_fullname}
    \bibliography{main}
}
\input{sec/X_supplementary}

\end{document}

%% file: preamble.tex
\usepackage[dvipsnames]{xcolor}


%% file: sec/0_abstract.tex
\begin{abstract}
Multi-instance registration is a challenging problem in computer vision and robotics, where multiple instances of an object need to be registered in a standard coordinate system. 
In this work, we propose the first iterative framework called instance-by-instance (IBI) for multi-instance 3D registration (MI-3DReg). 
It successively registers all instances in a given scenario, starting from the easiest and progressing to more challenging ones. 
Throughout the iterative process, outliers are eliminated continuously, leading to an increasing inlier rate for the remaining and more challenging instances.
Under the IBI framework, we further propose a sparse-to-dense-correspondence-based multi-instance registration method (IBI-S2DC) to achieve robust MI-3DReg. Experiments on the synthetic and real datasets have demonstrated the effectiveness of IBI and suggested the new state-of-the-art performance of IBI-S2DC, \eg, our MHF1 is 12.02\%/12.35\% higher than the existing state-of-the-art method ECC on the synthetic/real datasets.
\end{abstract}

%% file: sec/1_introduction.tex
\section{Introduction}
\label{sec: introduction}

Most research on 3D point cloud registration focuses on estimating a single transformation between pairwise point cloud~\cite{GORE, Graph-cut, DGR, DHVG, SACCOT, PointDSC, SC2, MAC}.
However, the target scene may contain multiple repeated instances in real applications, and the problem of estimating multiple transformations is called multi-instance 3D registration (MI-3DReg). 

This task has been relatively underexplored. Early works~\cite{PPF, PPFVoting} use point pair features (PPF) to detect and estimate the pose transformations of the instances. The quality of correspondence has been enhanced with the emergence of deep-learned features, resulting in performance boosting for correspondence-based MI-3DReg~\cite{ECC, PointCLM}. The first research on MI-3DReg based on correspondences is proposed in 2022~\cite{ECC}, \textit{i.e.}, efficient correspondence clustering (ECC), followed by a learned PointCLM method~\cite{PointCLM}. Existing methods generally register all instances in a one-shot manner, demonstrating their efficiency.
However, the correspondences between different instances can interfere with each other, making the registration of instances with low inlier rate very difficult. 

\begin{figure}
        \begin{subfigure}{0.6\linewidth}
		\centering
		\includegraphics[width=1\linewidth]{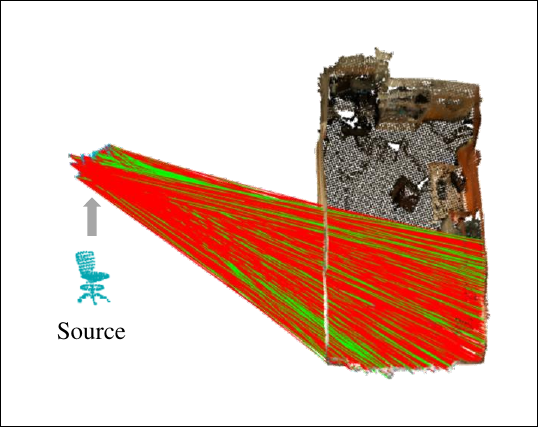}
  \caption{Input correspondences}
	\end{subfigure}
	\begin{subfigure}{0.36\linewidth}
		\centering
		\includegraphics[width=0.62\linewidth]{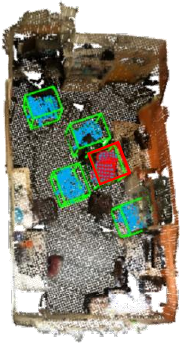}
  \caption{One-shot style (ECC)}
	\end{subfigure}
 \centering
         \begin{subfigure}{0.85\linewidth}
		\centering
		\includegraphics[width=1\linewidth]{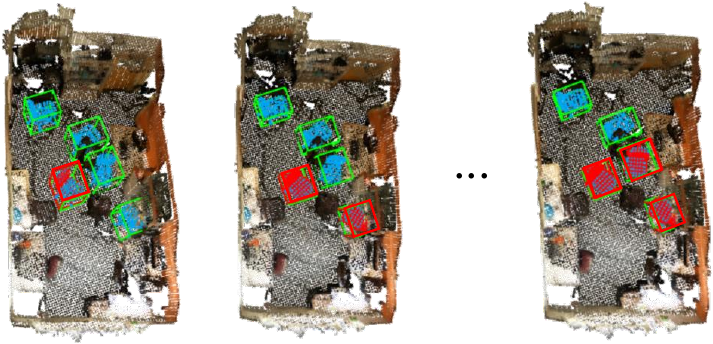}
  \caption{Iterative style (Our IBI)}
	\end{subfigure}
 \centering
	\caption{\label{fig:1}Comparison of one-shot method and our IBI for MI-3DReg. The inliers and outliers are visualized in green and red, respectively. The green and red bounding boxes represent the ground truth and estimated poses of instances, respectively.}
\centering
\end{figure}

To this end, we propose the first iterative framework called instance by instance (IBI) for robust MI-3DReg. 
We argue that inliers of a specific instance appear to be outliers to other instances. IBI iteratively registers instances and reduces outliers for each instance, which can effectively alleviate the registration problem for instances with scarce inliers.
First, a small set of highly consistent correspondences are selected as seed correspondences. 
Second, the seed correspondence set guides the enhancement of consistent correspondences to retrieve more inliers. 
Third, the six-degree-of-freedom (6-DoF) pose transformation is estimated with a guided sample consensus strategy. 
Finally, the generated registration hypothesis is globally validated, and the input correspondence set is updated by removing correspondences belonging to the current instance.
\cref{fig:1} shows that IBI is able to register more instances than traditional one-shot methods. 

The contributions are as follows:
    \begin{itemize}
        \item We propose IBI, the first iterative framework in MI-3DReg. It iteratively refines the registration results for each instance and gradually improves the precision and robustness of the registration process.
        \item Under the IBI framework, we introduce a sparse-to-dense-correspondence-based iterative method called IBI-S2DC. It mines consistent yet sparse seed correspondences to guide the retrieval of the whole inlier set. Its MHF1 is 12.02\%/12.35\% higher than the existing state-of-the-art method ECC~\cite{ECC} on the synthetic/real datasets.
    \end{itemize}

%% file: sec/2_related_work.tex
\section{Related Work}
\label{sec: related work}
\subsection{3D Point Cloud Registration}
\noindent
\textbf{Pairwise 3D point cloud registration.}
3D point cloud registration is an essential task in computer vision and robotics, including feature extraction, outlier rejection, and pose estimation.  
For the first stage, recent advances have highlighted the effectiveness of learning-based features over hand-designed ones in certain benchmarks, such as 3DMatch ~\cite{3DMatch} and SpinNet~\cite{SpinNet}. However, they still require robust outlier removal mechanisms. 
For the second stage, traditional outlier rejection methods are summarized by Yang \textit{et al.}~\cite{CorresGrouping}, including individual-based methods (\eg, SS~\cite{CorresGrouping}, NNSR~\cite{NNSR}, SI~\cite{SI}, CV~\cite{CV}) and group-based methods (\eg, RANSAC~\cite{RANSAC}, ST\cite{ST}, GC~\cite{GC}, 3DHV~\cite{3DHV}, GTM~\cite{GTM}). Other techniques, such as GORE~\cite{GORE} and PMC~\cite{PMC}, reduce outliers through geometric consistency tests. Bai \textit{et al.}~\cite{PointDSC} combined traditional methods and deep learning approaches to propose a two-stage network called PointDSC.
For the third stage, traditional methods such as RANSAC and its variants~\cite{RANSAC, Graph-cut, LnT, CG, SACCOT} follow a hypothesis verification process to estimate the 6-DoF pose transformation. On the other hand, FGR~\cite{FGR} and TEASER~\cite{TEASER} solve transformations directly from noisy correspondences using robust estimators.

\noindent
\textbf{Multi-instance 3D point cloud registration.} 
Multi-instance 3D point cloud registration is a complex problem in computer vision and robotics, involving the alignment of multiple object instances in a shared coordinate system. To the best of our knowledge, there are only two existing solutions for correspondence-based MI-3DReg. ECC~\cite{ECC} pioneers multi-instance point cloud registration by classifying initial correspondences into sets related to different instances, rejecting outliers, and resolving ambiguity. PointCLM~\cite{PointCLM}, a contrastive-learning-based method, presents representation learning and outlier pruning strategies suitable for MI-3DReg. Overall, the research particularly for MI-3DReg is still at an early stage.
\noindent
\subsection{Multi-model Fitting}
Theoretically, multi-model fitting methods can be also applied to MI-3DReg. Multi-model fitting methods estimate model parameters from data points generated by multiple models. 
Many existing methods perform one-shot model fitting. T-Linkage~\cite{T-Linkage} initializes a broad hypothesis set through point sampling and clustering with preference vectors for outlier removal. RansaCov~\cite{RansaCov} addresses multi-model fitting as a maximum coverage problem, offering approximate solutions. 
Alternatively, some methods rely on sequential RANSAC-based fitting. Progressive-X~\cite{Progressive-X} uses revised RANSAC with modified sampling weights to obtain distinct model parameters iteratively. CONSAC~\cite{CONSAC} incorporates deep models, akin to PointNet, for guiding the sampling process. However, they become inefficient when handling large-scale inputs.

Existing one-shot style MI-3DReg methods have overlooked the correspondence interference between instances, while multi-model fitting methods fail to incorporate proper point cloud geometric constraints. Motivated by these concerns, we propose a novel iterative IBI framework with two traits. First, it refines the registration results of each instance iteratively. Second, it enhances the registration accuracy and robustness progressively. 
It effectively alleviates the mutual interference issue caused by correspondences belonging to different instances, which is a common challenge encountered in one-shot methods.

%% file: sec/3_IBI_framework.tex
\section{The IBI Iterative Framework}
\label{sec: the IBI iterative framework}
\subsection{Problem Formulation}
For two point clouds $\mathbf{P}^{s}$ and $\mathbf{P}^{t}$ to be registered, $\mathbf{P}^{s}$ represents the source instance and there are multiple instances of $\mathbf{P}^{s}$ presented in $\mathbf{P}^{t}$. Let $\mathbf{p}^{s}$ and $\mathbf{p}^{t}$ respectively denote the points in the $\mathbf{P}^{s}$ and $\mathbf{P}^{t}$, IBI estimates the 6-DoF pose transformations $\{\mathbf{R}_{i},\mathbf{t}_{i}\}_{i=1,2,...,k}$ between the source instance in $\mathbf{P}^{s}$ and multiple instances in $\mathbf{P}^{t}$, where $k$ presents the number of repeated instances.
\begin{figure}[t]
\centering
\includegraphics[width=0.47\textwidth]{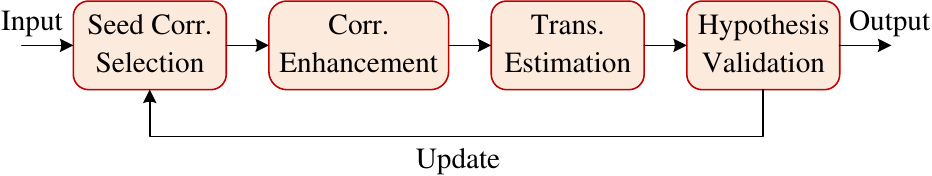}
\caption{\label{fig:2}Schematic illustration of the IBI framework.}
\end{figure}

\begin{figure*}[h]
\centering
\includegraphics[width=1.0\textwidth]{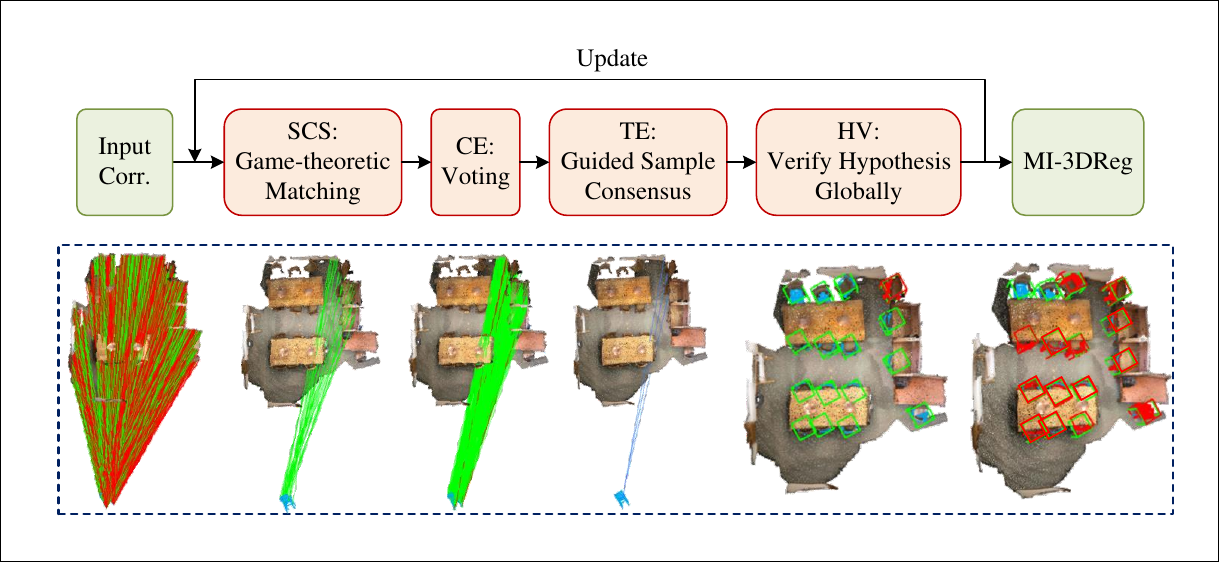}
\caption{\label{fig:3}The pipeline of IBI-S2DC. 1. SCS: Mine a sparse yet consistent correspondence set $\mathbf{C}_{s}$ under GTM guidance. 2. CE: Enhance consistency to achieve a dense correspondence set $\mathbf{C}_{d}$ based on a voting scheme. 3. TE: Estimate the transformation for the current instance based on a guided sample consensus estimator (GSAC). 4. HV: Validate the hypothesis globally using the point cloud overlap rate.}
\end{figure*}

\subsection{The Pipeline of IBI Framework}
An initial correspondence set $\mathbf{C} = \left\{\mathbf{c}\right\}$ is served as input, where $\mathbf{c} = (\mathbf{p}^{s}, \mathbf{p}^{t})$. It can be obtained by matching point cloud features.  For dense input, we usually downsample the $\mathbf{C}$ to $\mathbf{C}_{downsample}$, which contains $N_{downsample}$ correspondences. The IBI framework includes four steps, \textit{i.e.}, seed correspondence selection (SCS), correspondence enhancement (CE), transformation estimation (TE), and hypothesis validation (HV). First, a set of sparse yet consistent seed correspondences is selected to identify a single consistency from $\mathbf{C}_{downsample}$. Second, the seed correspondences are utilized to guide the enhancement of consistent correspondences. Third, the pose transformation of a single instance is estimated by a robust 6-DoF estimator. Finally, the generated registration hypotheses are validated, where a correct hypothesis indicates the identification of a valid instance in the target point cloud. After validation, the correspondences related to the identified instance are removed and the iterative framework proceeds to the next instance. The registration framework is illustrated in \cref{fig:2}.

%% file: sec/4_IBI-S2DC.tex
\section{A Trail on IBI for MI-3DReg}
\label{sec: A trail on IBI for MI-3DReg}
In this section, under the IBI framework, we propose IBI-S2DC for MI-3DReg, the pipeline is shown in \cref{fig:3}.

\subsection{SCS: Game-theoretic Matching}
\label{seed selection-game theory matching}
It is challenging to directly retrieve all inliers from the input because the scale is usually large and correspondences belonging to different instances exist simultaneously. To address this, we propose a sparse-to-dense-based mechanism. For sparse correspondences, we propose to mine a set of seed correspondences with high consistency under the game-theoretic matching (GTM)~\cite{PointCLM} guidance.

First, $\mathbf{P}^{s}$ and $\mathbf{P}^{t}$ are treated as a pair of players, and candidate correspondences in $\mathbf{C}_{downsample}$ are available strategies. The amount of population $\mathbf{x}$ that plays each strategy $\mathbf{c}_{i} $ at a given time is defined as:
\begin{equation}
\mathbf{x} = \left ( \mathbf{x}_{1},\dots,\mathbf{x}_{\left | \mathbf{C}_{downsample} \right | }    \right ) ^{T}.
\label{con:population}
\end{equation}

Then we set the initial population around the barycenter, and then the population will dynamically update with an evolutionary process by applying the following replicator dynamics equation: 

\begin{equation}
    \mathbf{x}_{i}\left ( t+1 \right )  = \mathbf{x}_{i}\left ( t \right )  \frac{\left (\Pi \mathbf{x}\left ( t \right )   \right )_{i}  }{\mathbf{x}\left ( t \right )^{T}\Pi \mathbf{x}\left ( t \right )   }, \label{con:GTMupdate}
\end{equation}
where $\Pi$ is the payoff matrix that assigns the payoff between strategies $\mathbf{c}_{i} $ 
and $\mathbf{c}_{j} $ to row $i$ and column $j$, and it is measured by the compatibility as:

\begin{equation}
    \Pi=\left\{ \begin{aligned}
  & r\left( \mathbf{c}_{i},\mathbf{c}_{j} \right),\quad \text{if} \quad \mathbf{c}_{i}\ne \mathbf{c}_{j}   \\
 & 0,\quad \text{otherwise}
\end{aligned}\right.\label{con:payoffmatrix}
,\end{equation}
where the rigidity term $r\left ( \mathbf{c}_{i},\mathbf{c}_{j}   \right )$ is:
\begin{equation}
    r\left ( \mathbf{c}_{i},\mathbf{c}_{j}   \right ) =\left | \left \| \mathbf{p}_{i}-\mathbf{p}_{j} \right \|-\left \| \mathbf{p}'_{i}-\mathbf{p}'_{j} \right \|   \right |. \label{con:rigidity}
\end{equation}

The dynamics will converge to a Nash equilibrium~\cite{Evolutionary}. After $N_{gtm}$ times of evolution, the correspondences whose strategies are greater than a threshold $t_{gtm} $, are served as inliers. This step makes an efficient selection of the input correspondence set $\mathbf{C}_{downsample}$, generating a sparse correspondence set $\mathbf{C}_{s}$, which is highly consistent.

\subsection{CE: Voting-based Enhancement}
\label{correspondences enhancement-consistency voting}
\begin{figure}
\centering
\includegraphics[width=0.46\textwidth]{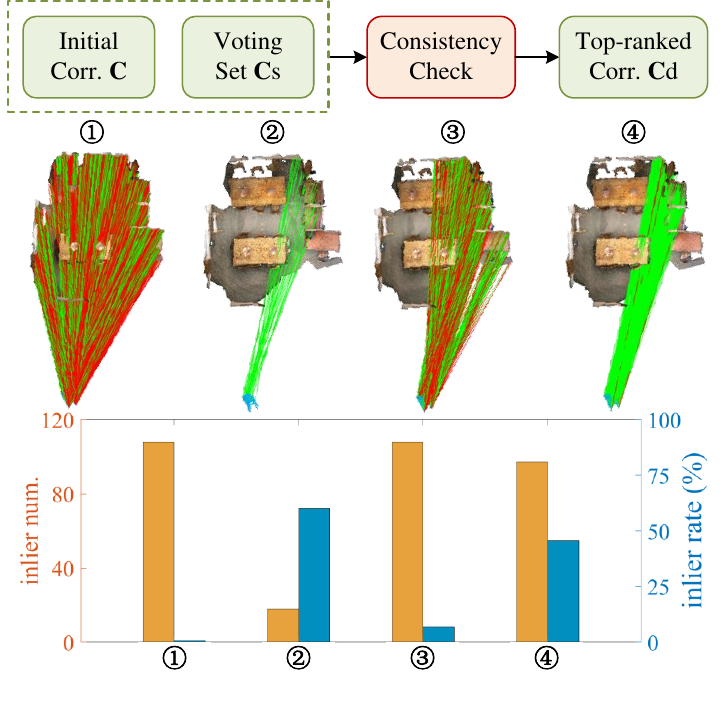}
\caption{\label{fig:4}The pipeline of the voting-based correspondence enhancement. After mining a sparse correspondence set, the inlier rate of the instance rises but the number of inliers is very small. The voting-based enhancement helps to retrieve significantly more inliers while maintaining a high inlier rate, resulting in a consistent and dense correspondence set.}
\end{figure}

After SCS, the seed correspondence set is consistent but sparse, which cannot guarantee the accuracy of instance registration. Therefore, we propose a voting-based method, in order to retrieve more inliers from the input correspondence set. In particular, it calculates a voting score for each correspondence based on the consistency check between the input correspondences and the voting set. We propose to serve the seed correspondence set as the voting set to improve the voting reliability. The pipeline of consistency voting is illustrated in \cref{fig:4}. 

First, the sparse correspondences set $\mathbf{C}_{s}$ is served as the voting set $\mathbf{C}_{vot}$. The correspondences in $\mathbf{C}_{s}$ are voters and those in $\mathbf{C}$ are candidates.

Second, we introduce a compatibility measure for a correspondence pair $\left ( \mathbf{c}_{i},\mathbf{c}_{j}   \right )$ with single consistency:
\begin{equation}
    D\left ( \mathbf{c}_{i},\mathbf{c}_{j}   \right ) =exp\left (-\frac{r\left ( \mathbf{c}_{i},\mathbf{c}_{j}   \right )^{2} }{\delta _{r}^{2} }\right ) ,\label{con:compatibility}
\end{equation}
where $\delta_{r}$ represents the rigidity parameters. 

Third, all voters cast their votes for the candidates in the input correspondence set $\mathbf{C}$. The final voting score of a correspondence $\mathbf{c}$ is defined as the sum of all compatibility scores of $\mathbf{c}$ and $\mathbf{c}_{i}\in \mathbf{C}_{vot}  $:

\begin{equation}
s\left( {\mathbf{c}} \right)=\sum\limits_{{\mathbf{c}_{i}\in \mathbf{C}_{vot}}}^{}{D\left ( {\mathbf{c},\mathbf{c}_{i}} \right)}.\label{con:votingscore}
\end{equation}
The correspondences are ranked based on voting scores in descending order, and the top $N_{vot}$ correspondences are kept as inliers. 

This step verifies and enhances the consistency of the sparse correspondence set $\mathbf{C}_{s}$, resulting in a dense correspondence set $\mathbf{C}_{d}$.

\subsection{TE: Guided Sample Consensus}
\label{guided sample consensus}
Single-instance 6-DoF pose transformation estimation is typically achieved with RANSAC~\cite{RANSAC}. However, the original RANSAC is sensitive to outliers and usually requires a number of iterations. 
In our TE step, we propose a guided sampling consensus (GSAC) method to fully utilize the voting score constraints as sampling weight. It can improve the accuracy of transformation estimation and reduce the number of iterations.

First, correspondence triplets are generated from $\mathbf{C}_{d}$ with three-point-sampling. Then, they are sorted in descending order based on the sum of the voting scores of the sampled three correspondences. After $N_{gsac}$ iterations, the generated hypotheses are evaluated to identify the optimal hypothesis as the output of GSAC. Specifically, we use the mean absolute error (MAE)~\cite{Metrics} to assess the quality of a hypothesis $\mathbf{T}_{i}$. The MAE score is defined as:

\begin{equation}\label{eq:mae}
S_{mae}\left( {\mathbf{T}_{i}} \right)=\sum\limits_{i=1}^{\left| \mathbf{C} \right|}{f\left( {{\mathbf{c}}_{i}} \right)},
\end{equation}
where the transformation error of $\mathbf{c}_{i}$ is defined as:

\begin{equation}
    f\left( \mathbf{c_{j}} \right)=\left\{ \begin{aligned}
  & \frac{\left| e\left( {{\mathbf{c}}_{i}} \right)-{{t}_{he}} \right|}{{{t}_{he}}},\quad \text{if} \quad e\left( {{\mathbf{c}}_{i}} \right)<{{t}_{he}} \\ 
 & 0,\quad \text{otherwise} \\ 
\end{aligned} \right.
,\end{equation}
where $e\left(\mathbf{c}_{i}\right)=\left\|\mathbf{R}_{i} \mathbf{p}_{i}^{s}+\mathbf{t}_{i}-\mathbf{p}_{i}^{t}\right\|$ represents the transformation error of $\mathbf{c}_{i}$. In addition, $t_{he}$ is a distance threshold to judge whether $\mathbf{c}_{i}$ is an inlier. The hypothesis with the highest MAE score is served as the final pose transformation for the current instance.

\subsection{HV: Verify Hypothesis Globally}
\label{verify hypothesis and update correspondences}
\begin{figure}[t]
\centering
\subcaptionbox{Correspondences with high inlier rate yet multiple consistencies}
    {%
        \includegraphics[width=0.96\linewidth]{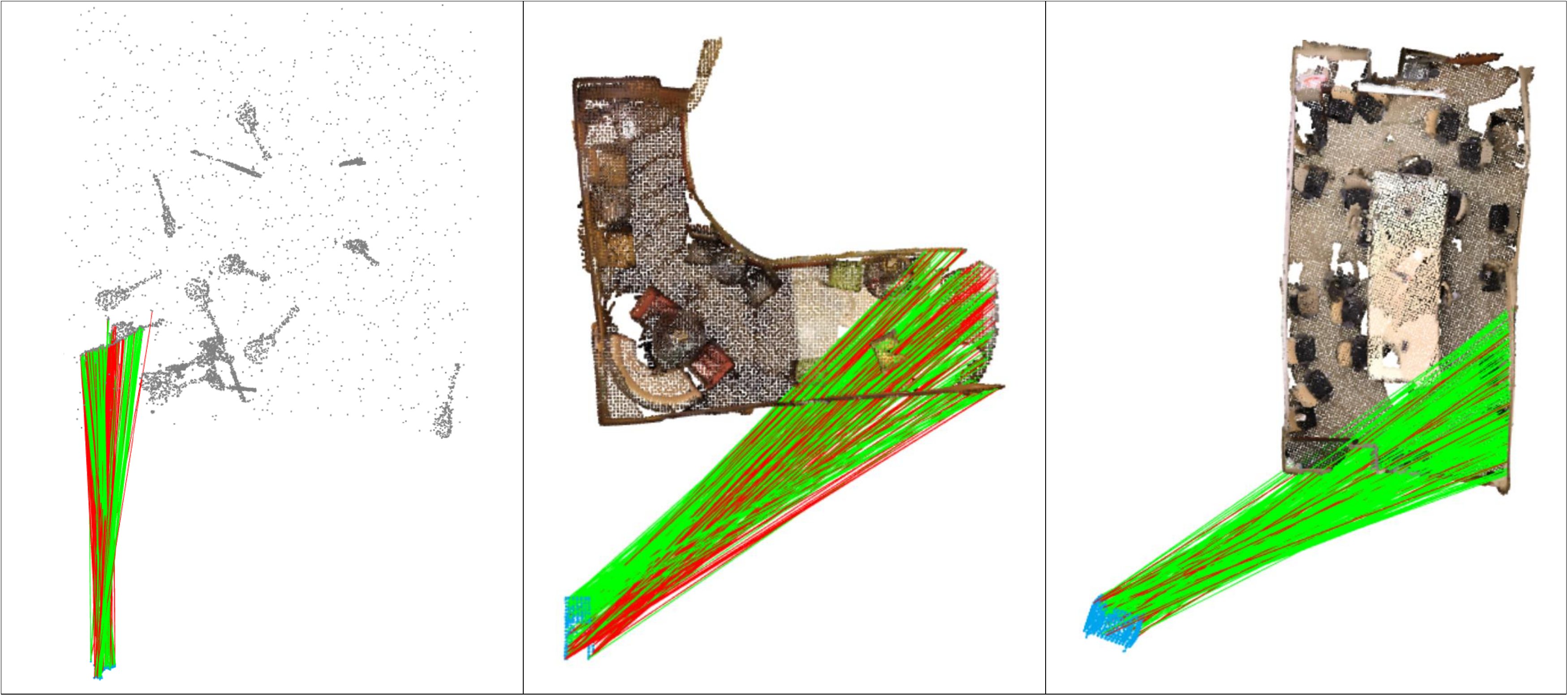}}
\subcaptionbox{Correspondences with low inlier rate and multiple consistencies}
    {%
        \includegraphics[width=0.96\linewidth]{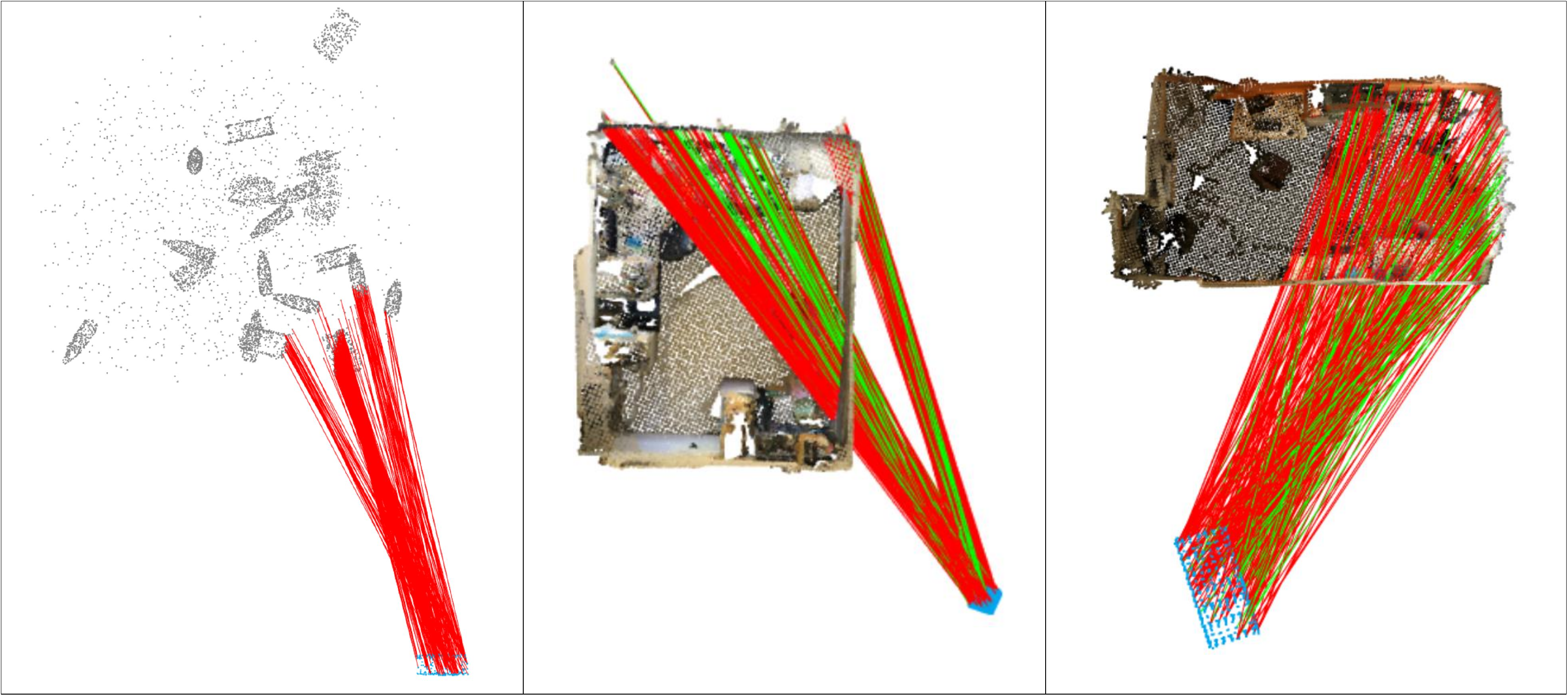}}
\caption{\label{fig:5}In challenging registration cases, correspondences may still exhibit multiple consistencies and suffer from outliers after the CE step. In (a), the dense correspondences have a high inlier rate, while multiple-consistencies exist. In (b), there are multiple consistencies and many outliers in the dense correspondence set.}
\end{figure}

After the previous three steps, we perform post-validation for the transformation generated by TE. The motivation is that correspondences between the scene and the source model, even after the CE step, may still exhibit multi-consistency and suffer from outliers, as shown in \cref{fig:5}. It is potentially due to that some instances are spatially very close, so the consistencies are hard to distinguish. This indicates that simply using correspondence-level information for hypothesis validation is not sufficient. Hence, we further propose a validation step that leverages global point-cloud-level information for hypothesis verification.

In particular, the overlap rate between two point clouds is considered for validation. First, we define the set of overlapped points as ~\cite{OP}:
\begin{equation}
    \mathbf{P}_{op}=\left\{{\mathbf{p}_{i}'^{s}}|d(\mathbf{p}_{i}'^{s},\mathbf{P}^{t})\leq d_{op}^{th},\mathbf{p}_{i}'^{s}\in\mathbf{P}'^{s}\right\},
\end{equation}
where $\mathbf{P}'^{s}$ represents the transformed point cloud, and the `point-to-surface' distance $d(\mathbf{p}_{i}'^{s},\mathbf{P}^{t})$ is:
\begin{equation}
d(\mathbf{p}_{i}'^{s},\mathbf{P}^{t})=\mathop{\min}_{\mathbf{p}_{j}^{t}\in\mathbf{P}^{t}}\left\|\mathbf{p}_{i}'^{s}-\mathbf{p}_{j}^{t}\right\|,
\end{equation}
where $d_{op}^{th}$ is a distance threshold to determine if two points are in the overlapped area of $\mathbf{P}^{s}$ and $\mathbf{P}^{t}$. 
Then we define the overlap rate as:
\begin{equation}
    overlap=\frac{|\mathbf{P}_{op}|}{|\mathbf{P}^{s}|},
\end{equation}
where $|\cdot|$ denotes the cardinality of a set.

If $overlap>t_{overlap}$, the hypothesis is accepted and an instance is supposed to be registered. Then its corresponding dense correspondence set $\mathbf{C}_{d}$ is removed from the initial correspondence set. This helps on one hand reduce the input size and, on the other, improve the inlier ratio of the remaining instances. Otherwise, the iteration will switch to the SCS step. Note that we still remove $\mathbf{C}_{d}$ when $overlap$ is smaller than $t_{overlap}$, to prevent the iteration from getting stuck in an endless loop.
Because the number of instances in the 3D scene is unknown, an iteration stop criterion is required. We assume that when the number of sparse correspondences $N_{s}<t_{s}$, the consistency is not strong enough, and the iteration stops (see Algorithm \ref{alg: IBI-S2DC algorithm} for pseudocode).
\begin{algorithm}[t]
    \caption{IBI-S2DC Algorithms}
    \label{alg: IBI-S2DC algorithm}
    \begin{algorithmic}[1] 
        \REQUIRE the point correspondences $\mathbf{C}$
        \ENSURE the 6-DoF pose transformations of target instances 
        \STATE {Downsample $\mathbf{C}$ to  $\mathbf{C}_{downsample}$ with $N_{downsample}$ correspondences}
        \WHILE {the size of $N_{s}>t_{s}$}
            \STATE {Initial the payoff matrix $\Pi(i,j)$}
            \FOR {$ i = 0 $ ; $ i < N_{gtm} $ ; $ i ++ $ }
                \STATE {Compute the population $\mathbf{V}(j)$}
                \STATE {Update $\mathbf{V}(j)$ dynamically by Eq.~\ref{con:GTMupdate}}
            \ENDFOR
            \STATE {Compute the sparse correspondence set $\mathbf{C}_{s}$}
            \FOR {$ j = 0 $ ; $ j < N_{vot}$ ; $ j ++ $ }
                \STATE {Compute consistency score by Eq. ~\ref{con:compatibility}}
                \STATE {Compute voting score by Eq. ~\ref{con:votingscore}}
                \STATE {Compute the dense correspondence set $\mathbf{C}_{d}$}
            \ENDFOR
            \STATE {Rank $\mathbf{C}_{d}$ by voting scores in descending order}
            \FOR {$ k = 0 $ ; $ k < N_{gsac}$ ; $ k ++ $ }
                \STATE {Perform guided three-point-sampling in $\mathbf{C}_{d}$}
                \STATE {Evaluate hypotheses with Eq.~\ref{eq:mae} and keep the best hypothesis}
            \ENDFOR
            \STATE {Validate the kept hypothesis}
            \IF {$overlap>t_{overlap}$}
                \STATE {Serve the hypothesis as the predicted transformation}
                \STATE {Update the input correspondence set $\mathbf{C}\leftarrow\mathbf{C}-\mathbf{C}_{d}$}
            \ENDIF
        \ENDWHILE
    \end{algorithmic} 
\end{algorithm}

%% file: sec/5_experiment.tex
\section{Experiment}
\label{sec:experiment}
\subsection{Experimental Setup}
\label{experiment setup}
\noindent
\textbf{Datasets.} We conduct experiments on both synthetic and real datasets. 
The synthetic dataset is created based on pre-sampling the ModelNet40 dataset using PointNet++~\cite{Pointnet++}. Each point cloud is downsampled to 256 points, and $K$ (up to 20) random transformations mixed with other objects and random points are applied to form the target point cloud.

Scan2CAD~\cite{Scan2CAD} serves as a real benchmark dataset aligning ShapeNet~\cite{ShapeNet} CAD models with object instances in ScanNet~\cite{Scannet} point clouds. We sample pairs that contain multiple CAD models in scans for testing.

\noindent
\textbf{Metrics.} We adopt the three metrics for MI-3DReg task proposed in ECC~\cite{ECC}, including Mean Hit Recall ($MHR$), Mean Hit Precision ($MHP$), and Mean Hit F1 ($MHF1$).

\noindent
\textbf{Implementation details.} We compare the proposed IBI-S2DC with five state-of-the-art methods for MI-3DReg: PointCLM (2022)~\cite{PointCLM}, ECC (2022)~\cite{ECC}, T-linkage (2014)~\cite{T-Linkage}, Progressive-X (2019)~\cite{Progressive-X}, and CONSAC (2020)~\cite{CONSAC}. For a fair comparison, all methods employ the same point correspondences as input. In PointCLM, we directly utilize the trained models in the synthetic and real datasets provided by the authors for evaluation.

The parameters of our method are set as follows: $N_{downsample}= 1024$, $N_{gtm}= 20$, and $N_{vot}=300$. The threshold $t_{gtm}$ is calculated by OSTU~\cite{OSTU}. The term $\delta_{r}$ in the CE step is set to 10 pr. Here, `pr' is the resolution of a point cloud~\cite{CV}. Stop critria $t_{s}$ is set to 5. For the synthetic dataset, we set $N_{gsac}=100$, $t_{he}=$10 pr, $d_{op}^{th}=$1.5 pr, and $t_{overlap}=0.85$. For the real dataset, we set $N_{gsac}=20$, $t_{he}=$1 pr, $d_{op}^{th}=$3 pr, and $t_{overlap}=0.7$. 

\subsection{Comparative Experiments}
\label{comparative experiments}
\subsubsection{Synthetic Dataset}

\begin{figure}[t]
        \begin{subfigure}{0.56\linewidth}
		\centering
		\includegraphics[width=1\linewidth]{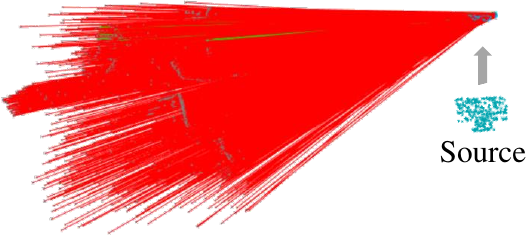}
  \caption{Input correspondences}
	\end{subfigure}
	\begin{subfigure}{0.4\linewidth}
		\centering
		\includegraphics[width=0.82\linewidth]{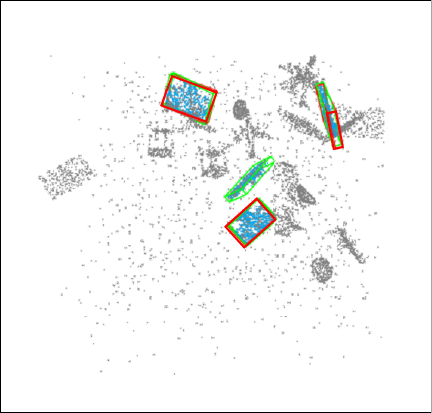}
  \caption{ECC (2022)}
	\end{subfigure}
 \centering
         \begin{subfigure}{0.48\linewidth}
		\centering
		\includegraphics[width=0.7\linewidth]{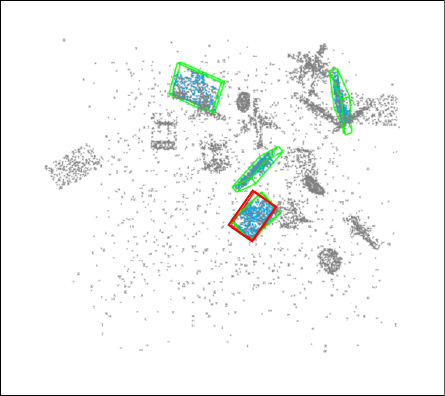}
  \caption{PointCLM (2022)}
        \end{subfigure}
        \begin{subfigure}{0.48\linewidth}
		\centering
		\includegraphics[width=0.7\linewidth]{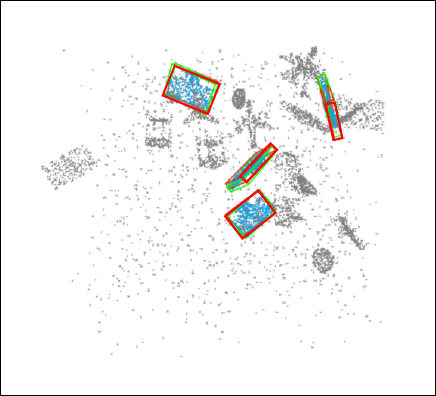}
  \caption{Ours}
	\end{subfigure}
 \centering
	\caption{\label{fig:6}Results on the synthetic dataset. (a) displays the input correspondences. (b-d) show the results of several compared methods. Here, estimated poses and the ground truth ones are rendered in red and green bounding boxes, respectively.}
\centering
\end{figure}

We employ the correspondences generated by PREDATOR~\cite{Predator}, as same as ECC~\cite{ECC}. Table \ref{tab1} shows the performance on the synthetic dataset. The results demonstrate that even under challenging initial conditions with an outlier rate exceeding 90\%, IBI-S2DC manages to perform admirably. The MHR is 61.16\%, the MHP is 71.20\%, and the MHF1 is 63.82\%. Although IBI-S2DC trails slightly behind PointCLM and ECC in terms of efficiency, its performance remains far better than those of the two competitors. We present some visual results in \cref{fig:6}. 

\begin{table}[t]
    \renewcommand{\arraystretch}{1.3}
	\centering
\resizebox{83mm}{!}{
\begin{tabular}{lllll}
\toprule
Metrics                     & MHR(\%) & MHP(\%) & MHF1(\%) & Times(s) \\
\midrule    
T-Linkage~\cite{T-Linkage}  &  0.19   &  0.54   &  0.27    & 43.46    \\
Prog-X~\cite{Progressive-X} & 15.90   & 31.01   & 18.98    & 86.39    \\
CONSAC~\cite{CONSAC}        &  0.10   &  0.07   &  0.08    &  7.65    \\
ECC~\cite{ECC}              & 53.39   & 61.44   & 51.80    &  1.28    \\
PointCLM~\cite{PointCLM}    &  0.62   &  0.18   &  0.28    & \bf{0.50}\\
\bf{IBI-S2DC}               &\bf{61.16}&\bf{71.20}&\bf{63.82}& 6.28  \\
\bottomrule
\end{tabular}}
\caption{Results on the synthetic dataset.}
\label{tab1}
\end{table}

\subsubsection{Real Dataset}

\begin{figure}[t]
        \begin{subfigure}{0.56\linewidth}
		\centering
		\includegraphics[width=1\linewidth]{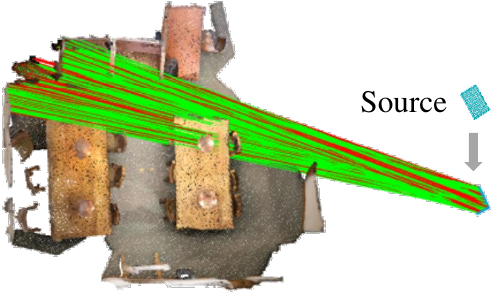}
  \caption{Input correspondences}
	\end{subfigure}
	\begin{subfigure}{0.4\linewidth}
		\centering
		\includegraphics[width=1\linewidth]{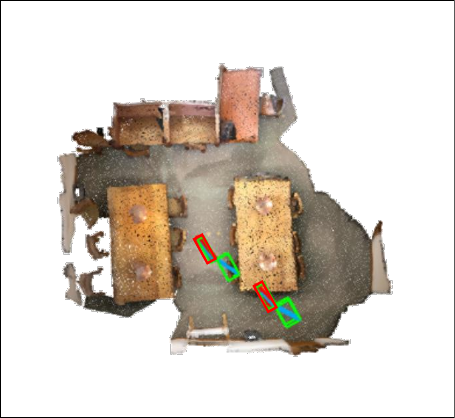}
  \caption{ECC (2022)}
	\end{subfigure}
 \centering
         \begin{subfigure}{0.48\linewidth}
		\centering
		\includegraphics[width=0.93\linewidth]{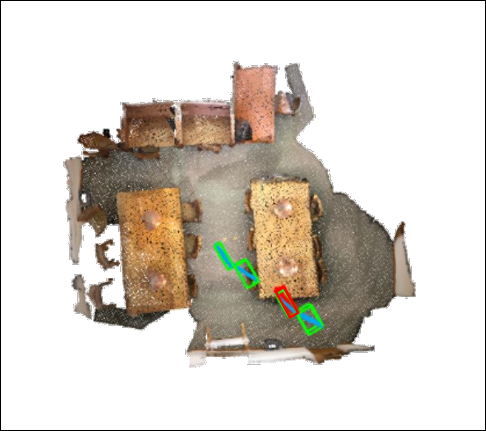}
  \caption{PointCLM (2022)}
        \end{subfigure}
        \begin{subfigure}{0.48\linewidth}
		\centering
		\includegraphics[width=0.94\linewidth]{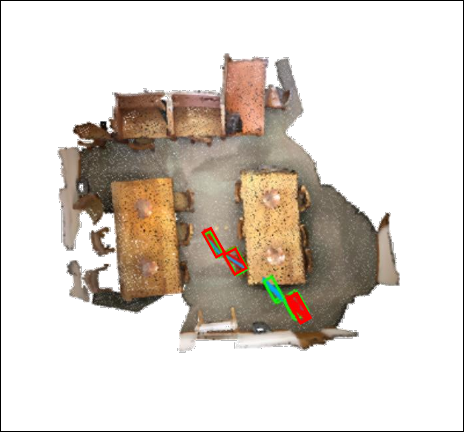}
  \caption{Ours}
	\end{subfigure}
 \centering
	\caption{\label{fig:7}Results on the real dataset.}
\centering
\end{figure}

Table \ref{tab2} reports the results on the real dataset. It indicates that for correspondence generated by PREDATOR~\cite{Predator} with an outlier rate exceeding 75\%, IBI-S2DC consistently outperforms all competitors. The MHR is 50.31\%, the MHP is 36.30\%, and the MHF1 is 39.39\%. This demonstrates the superiority of IBI-S2DC in large-scale and complex 3D real scenes. PointCLM, a deep-learned method, exhibits very limited performance when applied to the data setting by ECC.  \cref{fig:7} provides visualizations of several results.

\begin{table}[t]
    \renewcommand{\arraystretch}{1.3}
	\centering
\resizebox{83mm}{!}{
\begin{tabular}{lllll}
\toprule
Metrics                             & MHR(\%)   & MHP(\%)   & MHF1(\%)  & Times(s) \\
\midrule    
T-Linkage~\cite{T-Linkage}          &  2.46     &  3.79     &  2.71     & 1655.00    \\
Prog-X~\cite{Progressive-X}         & 11.58     &  6.86     &  7.87     & 26.32     \\
CONSAC~\cite{CONSAC}                &  2.66     &  0.35     &  0.62     & 21.35     \\
ECC~\cite{ECC}                      & 31.63     & 29.23     & 27.04     & 1.46     \\
PointCLM~\cite{PointCLM}            &  1.78     &  0.73     &  1.04     & \bf{0.19}    \\
\bf{IBI-S2DC}                       &\bf{50.31} &\bf{36.30} &\bf{39.39} & 7.56  \\
\bottomrule
\end{tabular}}
\caption{Results on the real dataset.}
\label{tab2}
\end{table}

\subsection{Analysis Experiments}
\label{analysis experiments}
\subsubsection{Validation Globally or Locally} 
We find that after the CE step, there may still be multiple consistencies in $\mathbf{C}_{d}$. Therefore, hypotheses need further validation. We compare the performance of validation globally (using point cloud overlap rate) and locally (using inlier count~\cite{Metrics}) to demonstrate our statement, as shown in Table \ref{tab3}. Two observations can be made. 1) Without hypothesis validation (referring to the control group), IBI-S2DC exhibits limited performance. 2) With global hypothesis validation, the result under different $t_{overlap}$ values consistently outperforms that with local hypothesis validation.
\begin{table}[t]
    \renewcommand{\arraystretch}{1.3}
	\centering
\resizebox{83mm}{!}{
\begin{tabular}{lllll}
\Xhline{0.8pt}
Metrics                             & MHR(\%) & MHP(\%) & MHF1(\%) & Times(s) \\
\hline    
\multicolumn{5}{l}{Control group: without hypothesis validation, $N_{s}<t_{s}$} \\
\hline
$t_{s} = 5$       & 59.56  & 26.03  & 32.68  & 6.21     \\
$t_{s} = 10$      & 54.90  & 26.46  & 32.03  & 6.03     \\
$t_{s} = 15$      & 50.05  & 26.97  & 31.33  & 5.24     \\
\hline   
\multicolumn{5}{l}{Experimental group: the inlier count, $\#inliers>t_{inliers}$} \\
\hline
$t_{inliers} = 50$     & 60.12  & 25.15  & 32.03  & 8.02     \\
$t_{inliers} = 100$    & 51.25  & 25.69  & 30.56  & 7.60     \\
$t_{inliers} = 150$    & 28.29  & 26.04  & 23.56  & 3.97     \\
\hline    
\multicolumn{5}{l}{Experimental group: point cloud overlap rate, $overlap>t_{overlap}$} \\
\hline 
$t_{overlap} = 0.8$    & 60.50  & 66.84  & 61.55  & 5.48     \\
$t_{overlap} = 0.85$   & 61.16  & 71.20  & 63.82  & 6.28     \\
$t_{overlap} = 0.9$    & 58.34  & 76.60  & 64.35  & 8.28     \\
\Xhline{0.8pt}
\end{tabular}}
\caption{The performance of IBI-S2DC with different hypothesis validation methods on the synthetic dataset.}
\label{tab3}
\end{table}

\subsubsection{Robustness Analysis}
\begin{figure}[t]
        \begin{subfigure}{0.48\linewidth}
		\centering
		\includegraphics[width=1\linewidth]{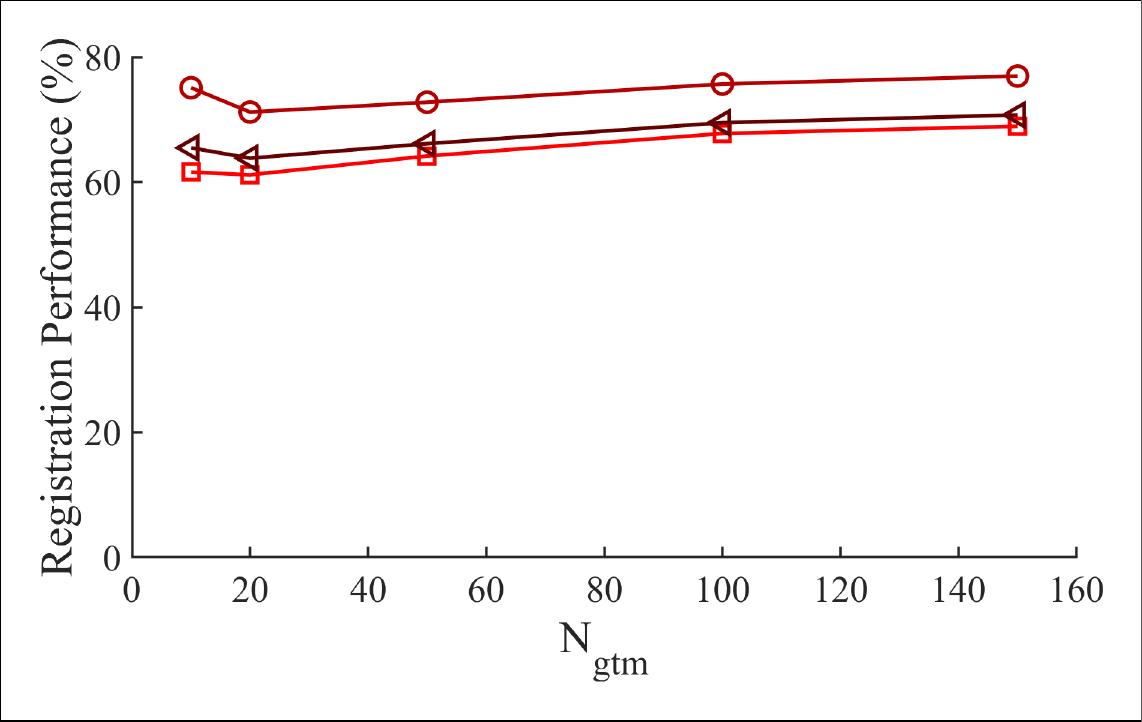}
  \caption{Synthetic dataset}
	\end{subfigure}
	\begin{subfigure}{0.48\linewidth}
		\centering
		\includegraphics[width=1\linewidth]{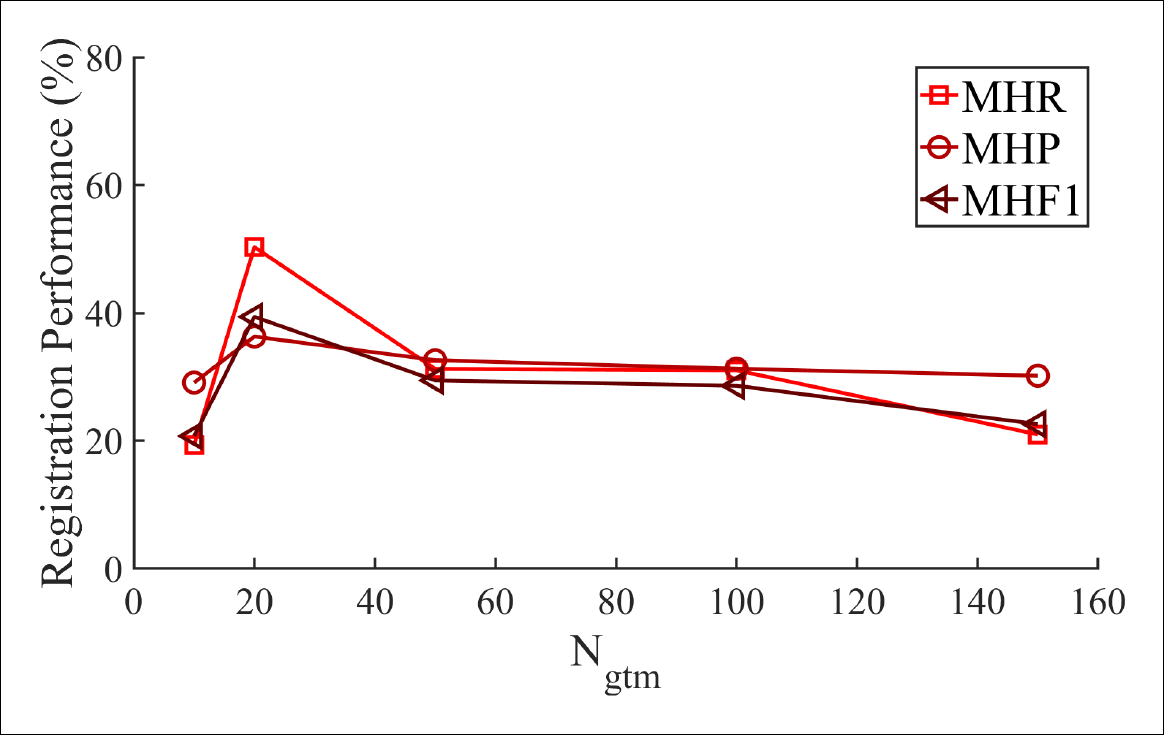}
  \caption{Real dataset}
	\end{subfigure}
 \centering
	\caption{\label{fig:8}Robustness to $N_{gtm}$ variation on the synthetic and real datasets.}
\centering
\end{figure}

\noindent
\textbf{Iterations of GTM.} We analyze the impact of $N_{gtm}$ in the SCS step, and set $N_{vot} = 300$ and $t_{s} = 5$. \cref{fig:8} shows the performance of our method under different $N_{gtm}$ values. The results suggest that our method is very robust to $N_{gtm}$ changes on the synthetic dataset. On the real dataset, the performance fluctuates slightly when $N_{gtm}$ varies from 10 to 50. Overall, our method is robust to $N_{gtm}$ variation.

\begin{figure}[t]
        \begin{subfigure}{0.48\linewidth}
		\centering
		\includegraphics[width=1\linewidth]{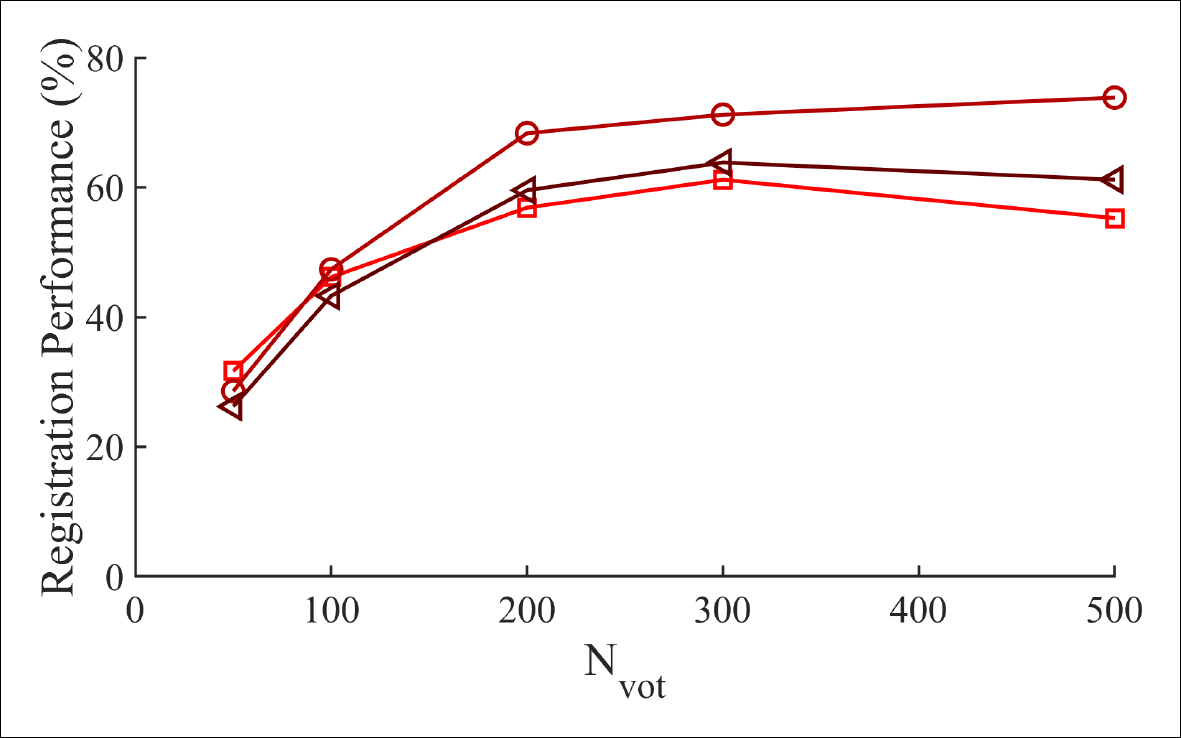}
  \caption{Synthetic dataset}
	\end{subfigure}
	\begin{subfigure}{0.48\linewidth}
		\centering
		\includegraphics[width=1\linewidth]{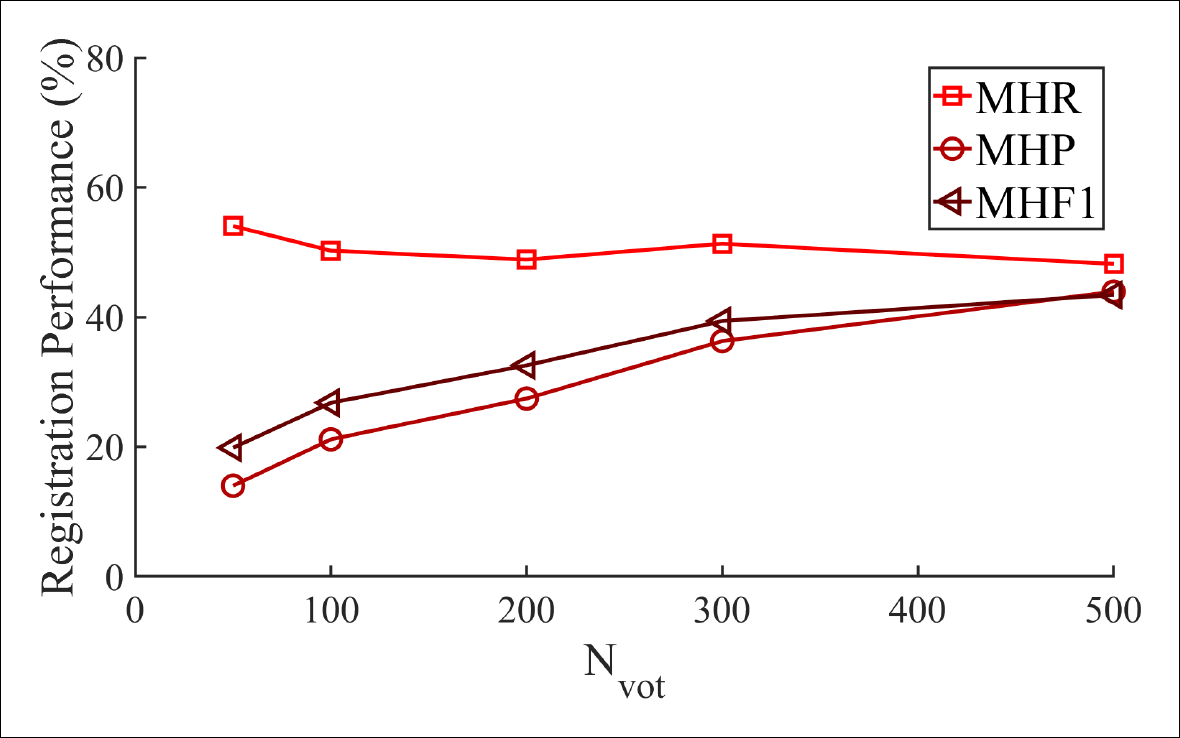}
  \caption{Real dataset}
	\end{subfigure}
 \centering
	\caption{\label{fig:9}Robustness to $N_{vot}$ variation on the synthetic and real datasets.}
\centering
\end{figure}

\noindent
\textbf{Number of correspondences after voting.}  Here, we set $N_{gtm} = 20$, $t_{s} = 5$ and vary $N_{vot}$. \cref{fig:9} shows the performance of our method under different $N_{vot}$ values. Generally, larger voting sets result in better performance. This indicates that more inliers are contained, while more consistency checks are required. When $N_{vot}$ is greater than 200, the performance is very stable.

\begin{figure}[t]
        \begin{subfigure}{0.48\linewidth}
		\centering
		\includegraphics[width=1\linewidth]{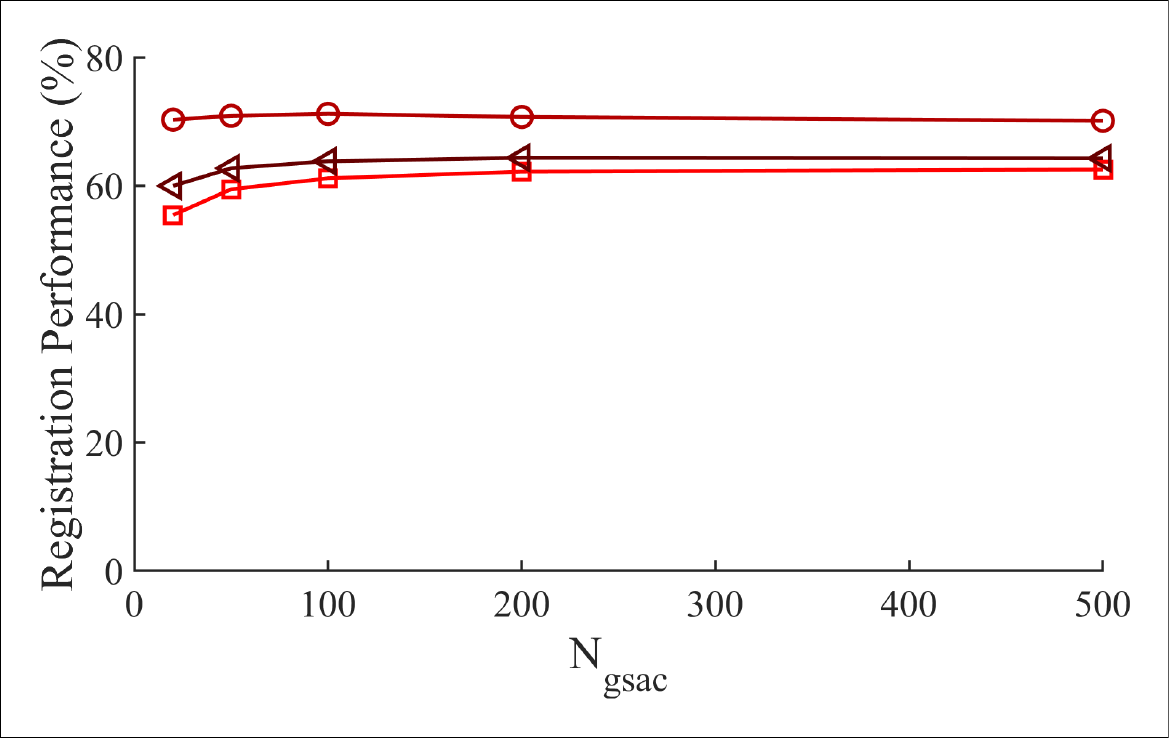}
  \caption{Synthetic dataset}
	\end{subfigure}
	\begin{subfigure}{0.48\linewidth}
		\centering
		\includegraphics[width=1\linewidth]{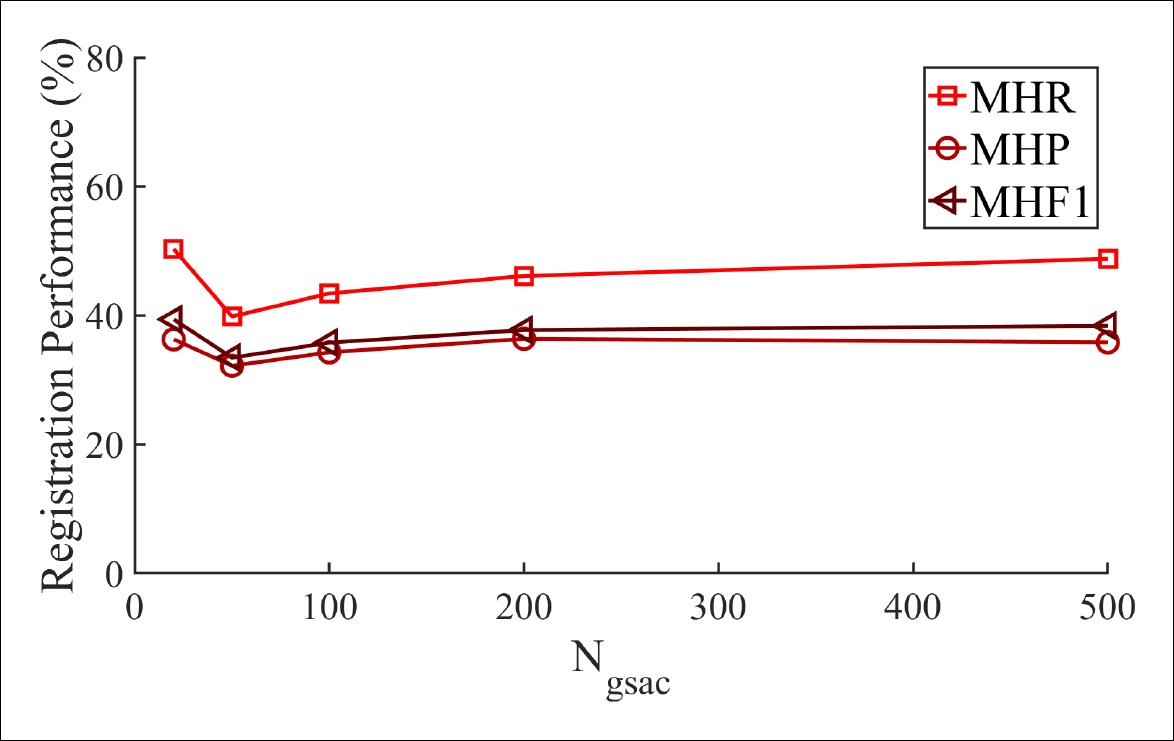}
  \caption{Real dataset}
	\end{subfigure}
 \centering
	\caption{\label{fig:10}Robustness to $N_{gsac}$ variation on the synthetic and real datasets.}
\centering
\end{figure}

\noindent
\textbf{Iterations of GSAC.} We set $N_{gtm} = 20$, $N_{vot} = 300$, $t_{s} = 5$, and vary $N_{gsac}$ in this experiment. \cref{fig:10} shows the performance of our method under different $N_{gsac}$ values. The results demonstrate that IBI-S2DC is robust to the $N_{gsac}$ setting. Even with 20 iterations, we are able achieve accurate 3D registration. 

\noindent
\textbf{Varying correspondence densities.} We examine the robustness of IBI-S2DC in the presence of inputs with various densities. Here, we set $N_{gtm} = 20$, $N_{vot} = 300$, $t_{s} = 5$, and $N_{gsac}$ to 100 and 20 on the synthetic and real datasets, respectively. Table \ref{tab4} and Table \ref{tab5} report the results. On the synthetic dataset, our IBI-S2DC consistently outperforms ECC. In addition, we find that IBI-S2DC is quite stable when the number of input correspondences varies from 512 to 2048. When faced with extremely sparse input (the correspondence count is smaller than 512), IBI-S2DC also yields limited performance. On the real dataset, IBI-S2DC consistently exhibits superior performance. Meanwhile, IBI-S2DC also achieves good performance with 256 input correspondences on the real dataset. These results indicate that IBI-S2DC possesses strong robustness to varying correspondence densities.

\begin{table}[t]
    \renewcommand{\arraystretch}{1.3}
	\centering
\resizebox{83mm}{!}{
\begin{tabular}{lll|llll}
\Xhline{0.8pt}
Metrics &    &    & MHR(\%)  & MHP(\%)  & MHF1(\%)  & Times(s)\\
\hline
\multicolumn{2}{l}{$N_{downsample}$} &  &  &  &  &  \\
\hline    
\multirow{2}*{256}  &ECC~\cite{ECC} &   &\quad -&\quad -&\quad -&\quad - \\  
~                   &IBI-S2DC       &   &  6.71 & 31.91 &  9.69 & 0.88   \\
\hline  
\multirow{2}*{512}  &ECC~\cite{ECC} &   & 48.78 & 62.92 & 51.26 & \bf{0.32}\\
~                   &IBI-S2DC       &   & \bf{59.66} & \bf{72.25} & \bf{63.42} & 6.11   \\ 
\hline  
\multirow{2}*{1024} &ECC~\cite{ECC} &   & 53.39 & 61.44 & 51.80 & \bf{1.28}   \\
~                   &IBI-S2DC       &   & \bf{61.16} & \bf{71.20} & \bf{63.82} & 6.28   \\ 
\hline 
\multirow{2}*{2048} &ECC~\cite{ECC} &   & 53.98 & 59.37 & 50.41 & \bf{2.05}   \\
~                   &IBI-S2DC       &   & \bf{54.91} & \bf{72.92} & \bf{60.32} & 6.61   \\  
\hline 
\multirow{2}*{5096} &ECC~\cite{ECC} &   &\quad -&\quad -&\quad -&\quad - \\
~                   &IBI-S2DC       &   & 47.23 & 71.52 & 54.14 & 8.62   \\ 
\Xhline{0.8pt}
\end{tabular}}
\caption{Performance with different $N_{downsample}$ values on the synthetic dataset. The term `-' represents that it fails to get results under the tested scenario.}
\label{tab4}
\end{table}

\begin{table}[t]
    \renewcommand{\arraystretch}{1.3}
	\centering
\resizebox{83mm}{!}{
\begin{tabular}{lll|llll}
\Xhline{0.8pt}
Metrics &    &    & MHR(\%)  & MHP(\%)  & MHF1(\%)  & Times(s)\\
\hline
\multicolumn{2}{l}{$N_{downsample}$} &  &  &  &  &  \\
\hline    
\multirow{2}*{256}  &ECC~\cite{ECC} &   & 32.11 & 28.70 & 28.40 &  \bf{0.32}  \\  
~                   &IBI-S2DC       &   & \bf{37.36} & \bf{36.67} & \bf{32.43} &  3.58  \\
\hline  
\multirow{2}*{512}  &ECC~\cite{ECC} &   & 32.18 & 29.11 & 27.57 &  \bf{0.61}  \\
~                   &IBI-S2DC       &   & \bf{44.76} & \bf{33.70} & \bf{35.98} &  4.24  \\  
\hline  
\multirow{2}*{1024} &ECC~\cite{ECC} &   & 31.63 & 29.23 & 27.04 & \bf{1.46}  \\
~                   &IBI-S2DC       &   & \bf{50.31} & \bf{36.30} & \bf{39.39}  & 7.56   \\  
\hline 
\multirow{2}*{2048} &ECC~\cite{ECC} &   &\quad -&\quad -&\quad -&\quad - \\
~                   &IBI-S2DC       &   & 50.91 & 37.31 & 40.50 &  6.99  \\   
\hline    
\multirow{2}*{5096} &ECC~\cite{ECC} &   &\quad -&\quad -&\quad -&\quad - \\
~                   &IBI-S2DC       &   & 50.82 & 37.60 & 40.30 &  9.26  \\
\Xhline{0.8pt}
\end{tabular}}
\caption{Performance with different $N_{downsample}$ values on the real dataset.}
\label{tab5}
\end{table}

\subsubsection{Module Effectiveness Analysis}
Table \ref{tab6} and Table \ref{tab7} present the effectiveness analysis on the four modules of our IBI method, \ie, SCS, CE, TE, and HV, on the synthetic and real datasets.

First, we analyze the SCS module. We remove the SCS module and replace the voting set in CE with the top-ranked correspondences sorted by nearest neighbor similarity ratio (NNSR)~\cite{NNSR}. Without SCS, the MHF1 of IBI-S2DC decreases 52.32\% and 10.85\% on the synthetic and real datasets, respectively. This verifies the necessity of mining consistency with a sparse set. 3D Hough voting~\cite{3DHV} is conducted as an alternative method.

Second, we ablate the CE module. We set the population of  
$\mathbf{C}_{s}$ as the guided sampling score for GSAC. Applying CE achieves MHF1 improvements of 37.29\% and 21.77\% on the synthetic and real datasets, respectively. This demonstrates that the CE module is critical to retrieving inliers and achieving accurate registration.

Third, we analyze the TE module. We replace GSAC with RANSAC, SACCOT, and MAC. Compared to GSAC, other solvers boost efficiency notably but meet a clear performance deterioration. This suggests that GSAC fully leverages the consistency information of the SCS and CE modules and effectively improves registration accuracy.

Fourth, we remove the HV module. Without hypothesis validation, the MHF1 results on the synthetic and real datasets drop by 31.31\% and 23.45\%, respectively. Local validation in the real dataset achieves better performance. 

Finally, \textit{we have also demonstrated that IBI is a general framework}. For instance, when employing RANSAC in the TE step, it achieves 59.69\% in terms of MHF1, with a reduced time consumption at 0.63 seconds. This variant of IBI still surpasses ECC, demonstrating the flexibility of IBI. We believe the performance of IBI can be further improved with more advanced methods for each module.

\begin{table}[t]
    \renewcommand{\arraystretch}{1.3}
	\centering
\resizebox{83mm}{!}{
\begin{tabular}{llllllll}
\toprule
SCS & CE & TE & HV & MHR(\%) & MHP(\%) & MHF1(\%) & Times(s) \\
\midrule    
\multicolumn{1}{c}{N}&      &      &      & 35.76 &  7.50 & 11.50 & 7.56   \\
\multicolumn{1}{c}{H}&      &      &      & 44.26 & \bf{72.62} & 52.51 & 66.65   \\
      &\multicolumn{1}{c}{\XSolidBrush}&      &      & 25.50 & 34.02 & 26.53 & \bf{0.38}   \\
      &      &\multicolumn{1}{c}{R}           &      & 54.25 & \underline{72.61} & 59.69 & \underline{0.63}   \\
      &      &\multicolumn{1}{c}{S}           &      & 57.07 & 70.13 & \underline{60.68} & 304.75   \\
      &      &\multicolumn{1}{c}{M}           &      & 57.56 & 68.13 & 60.00 & 12.04   \\
      &      &      &\multicolumn{1}{c}{\XSolidBrush}& \underline{60.44} & 25.70 & 32.51 & 6.35   \\
      &      &      &\multicolumn{1}{c}{L}& 49.57 & 69.30 & 54.32 & 7.51 \\
\multicolumn{1}{c}{\Checkmark} &\multicolumn{1}{c}{\Checkmark} &\multicolumn{1}{c}{\Checkmark} &\multicolumn{1}{c}{\Checkmark}     & \bf{61.16} & 71.20 & \bf{63.82} & 6.28\\
\bottomrule
\end{tabular}}
\caption{Module effectiveness analysis on the synthetic dataset. `N' represents NNSR, `H' represents 3D Hough Voting~\cite{3DHV}, `R' represents RANSAC\cite{RANSAC}, `S' represents SAC-COT~\cite{SACCOT}, `M' represents MAC~\cite{MAC} and `L' represents validate locally~\cite{Metrics}.}
\label{tab6}
\end{table}

\begin{table}[t]
    \renewcommand{\arraystretch}{1.3}
	\centering
\resizebox{83mm}{!}{
\begin{tabular}{llllllll}
\toprule
SCS & CE & TE & HV & MHR(\%) & MHP(\%) & MHF1(\%) & Times(s) \\
\midrule    
\multicolumn{1}{c}{N}&      &      &      & 34.47 & 28.92 & 28.55 &  6.41  \\
\multicolumn{1}{c}{H}&      &      &      & 33.18 & 31.02 & 29.69 & 46.23   \\
      &\multicolumn{1}{c}{\XSolidBrush}&      &      & 35.31 & 15.41 & 17.62 & \bf{0.30}   \\
      &      &\multicolumn{1}{c}{R}           &      & 29.30 & 28.28 & 26.17 & \underline{0.52}   \\
      &      &\multicolumn{1}{c}{S}           &      & \underline{57.73} & 36.14 & \underline{41.28} & 291.53 \\
      &      &\multicolumn{1}{c}{M}           &      & \bf{60.05} & 34.36 & 40.28 & 22.46 \\
      &      &      &\multicolumn{1}{c}{\XSolidBrush}& 28.04 & 12.58 & 15.94 & 3.80  \\
      &      &      &\multicolumn{1}{c}{L}& 52.59 & \bf{51.72} & \bf{48.54} & 6.47  \\
\multicolumn{1}{c}{\Checkmark} &\multicolumn{1}{c}{\Checkmark} &\multicolumn{1}{c}{\Checkmark}  &\multicolumn{1}{c}{\Checkmark}     &  50.31 & \underline{36.30} & 39.39 &  7.56  \\
\bottomrule
\end{tabular}}
\caption{Module effectiveness analysis on the real dataset.}
\label{tab7}
\end{table}

%% file: sec/6_conclusion.tex
\section{Conclusion}
\label{sec:conclusion}
In this paper, we presented an IBI framework for MI-3DReg by registering instances and reducing outliers iteratively. Under the framework, we further proposed IBI-S2DC and achieved state-of-the-art performance on all tested datasets. In particular, its MHF1 values are 12.02\%/12.35\% higher than the existing state-of-the-art method ECC on the synthetic/real datasets. We expect more trials under the proposed IBI framework in the future.

%% file: sec/X_supplementary.tex
\clearpage
\setcounter{page}{1}

\section{Performance with Varying Outlier Ratios}
In this section, we create input correspondences by mixing GT correspondences with outliers instead of generating correspondences by PREDATOR~\cite{Predator}, in order to control the outlier ratio $Outlier\_ratio$. Various outlier ratios were examined, including $10\%\sim50\%$, $50\%\sim70\%$, and $70\%\sim90\%$. Outliers were randomly selected within specified ranges for each test instance. For parameters, we set $N_{gtm} = 20$, $N_{vot} = 300$, $N_{gsac} = 100$, and $t_{select} = 5$. The outcomes are detailed in Table \ref{tab8}. Our IBI-S2DC achieves notable and stable performance under each range of $Outlier\_ratio$, and its MHF1 surpasses 90\% under $10\%\sim90\%$ outlier ratio. Our MHF1 also achieves 88.51\% when facing extreme situations, \textit{i.e.}, the outlier ratio exceeds 90\%. This clearly indicates the robustness of our IBI-S2DC to varying outlier ratios.

\begin{table}[h]
    \renewcommand{\arraystretch}{1.3}
	\centering
\resizebox{83mm}{!}{
\begin{tabular}{lll|llll}
\Xhline{0.8pt}
Metrics &    &    & MHR(\%)  & MHP(\%)  & MHF1(\%)  & Times(s)\\
\hline
\multicolumn{2}{l}{$Outlier\_ratio$} &  &  &  &  &  \\
\hline    
\multirow{2}*{$10\%\sim50\%$}&ECC~\cite{ECC} &   & 21.72 & 61.90 & 29.70 &\bf{1.58}\\ 
~             &IBI-S2DC      &   & \bf{97.74} & \bf{88.10} & \bf{92.36} & 2.85\\
\hline  
\multirow{2}*{$50\%\sim70\%$}&ECC~\cite{ECC} &   & 21.01 & 53.49 & 27.87 & \bf{2.22}\\
~             &IBI-S2DC      &   & \bf{98.41} & \bf{85.79} & \bf{91.30} & 4.04\\ 
\hline  
\multirow{2}*{$70\%\sim90\%$}&ECC~\cite{ECC} &   & 20.65 & 33.41 & 22.99 & \bf{2.60}\\
~             &IBI-S2DC      &   & \bf{98.49} & \bf{84.05} & \bf{90.25} & 5.39\\ 
\hline 
\multirow{2}*{$90\%\sim99\%$}&ECC~\cite{ECC} &   &\quad -&\quad -&\quad -&\quad -   \\
~             &IBI-S2DC      &   & 99.10 & 80.94 & 88.51 & 13.62  \\  
\Xhline{0.8pt}
\end{tabular}}
\caption{Performance with different $Outlier\_ratio$ values on the synthetic dataset.}
\label{tab8}
\end{table}

\section{Comparison with Correspondence-free Methods}
Although our main focus is on correspondence-based methods for MI-3DReg, there are also some correspondence-free methods, typically including point-pair-feature-based (PPF) ones. Here, two PPF methods for comparison, including Drost\_PPF~\cite{PPF} and Central\_Voting\_PPF~\cite{PPFVoting}. Table \ref{tab9} and Table \ref{tab10} present the comparative results. Clearly, both PPF methods achieve very limited performance, and our IBI-S2DC achieves the best performance. This is because stable PPFs usually rely on dense and high-quality point clouds, while the datasets used for MI-3DReg are sparse and noisy.
\begin{table}[h]
    \renewcommand{\arraystretch}{1.3}
	\centering
\resizebox{83mm}{!}{
\begin{tabular}{lllll}
\toprule
Metrics                              & MHR(\%) & MHP(\%) & MHF1(\%) & Times(s) \\
\midrule    
Drost\_PPF~\cite{PPF}                &  7.52    &  5.45    &  5.34    &\bf{2.20}\\
Central\_Voting\_PPF~\cite{PPFVoting}&  3.61    &  3.78    &  3.15    &  5.68  \\
IBI-S2DC                             &\bf{61.16}&\bf{71.20}&\bf{63.82}& 6.28   \\
\bottomrule
\end{tabular}}
\caption{Comparison with PPF methods on the synthetic dataset.}
\label{tab9}
\end{table}

\begin{table}[h]
    \renewcommand{\arraystretch}{1.3}
	\centering
\resizebox{83mm}{!}{
\begin{tabular}{lllll}
\toprule
Metrics                             & MHR(\%)   & MHP(\%)   & MHF1(\%)  & Times(s) \\
\midrule    
Drost\_PPF~\cite{PPF}                &  1.70    &  1.64    &  1.64    & 2.64   \\
Central\_Voting\_PPF~\cite{PPFVoting}&  0.45    &  0.40    &  0.40    &\bf{1.14}\\
IBI-S2DC                             &\bf{50.31}&\bf{36.30}&\bf{39.39}& 7.56   \\
\bottomrule
\end{tabular}}
\caption{Comparison with PPF methods on the real dataset.}
\label{tab10}
\end{table}
\section{More Visualizations}
\begin{figure*}
\centering
\includegraphics[width=1\textwidth]{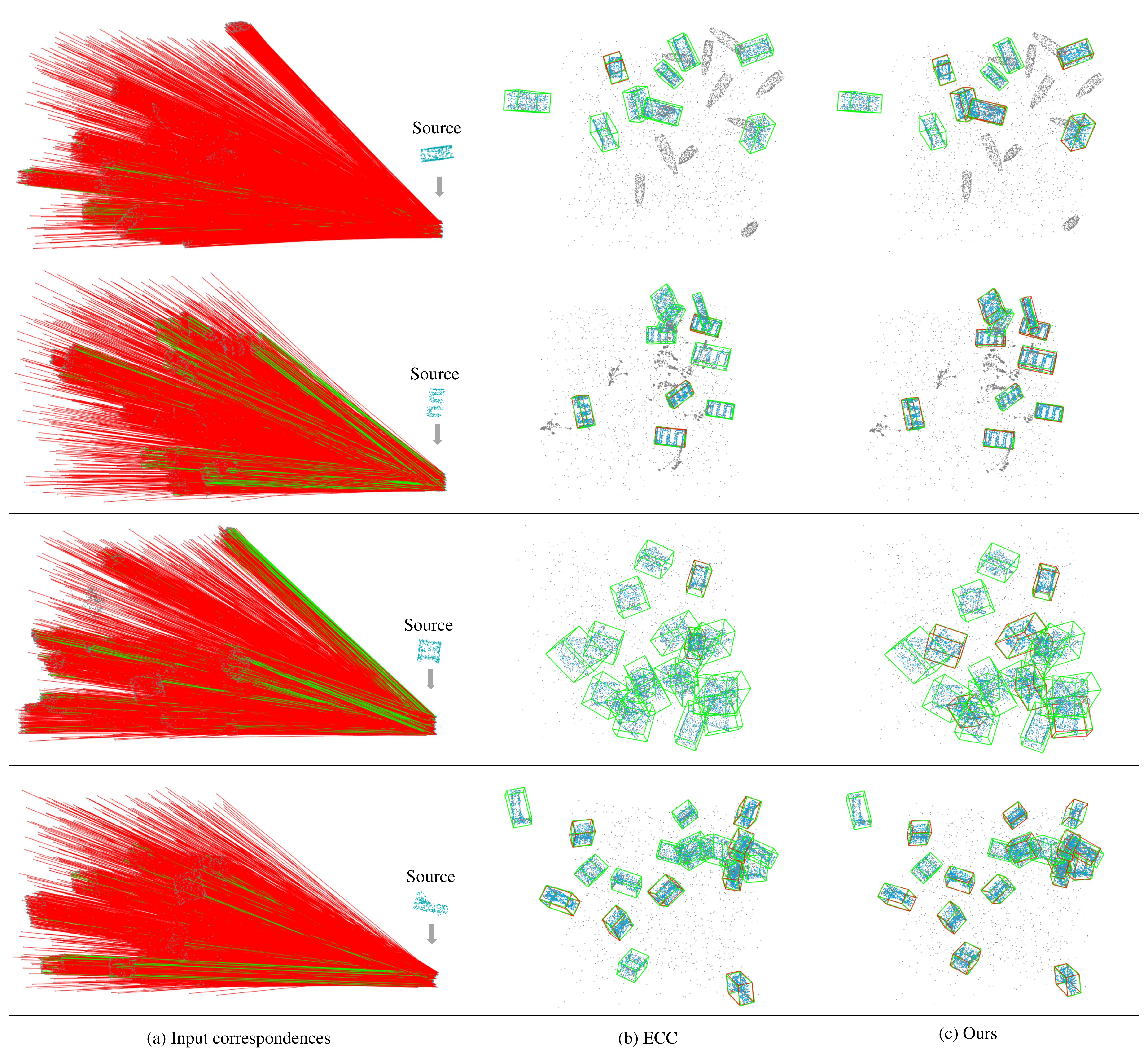}
\caption{\label{fig:11}Visual results on the synthetic dataset.}
\end{figure*}

\begin{figure*}
\centering
\includegraphics[width=1\textwidth]{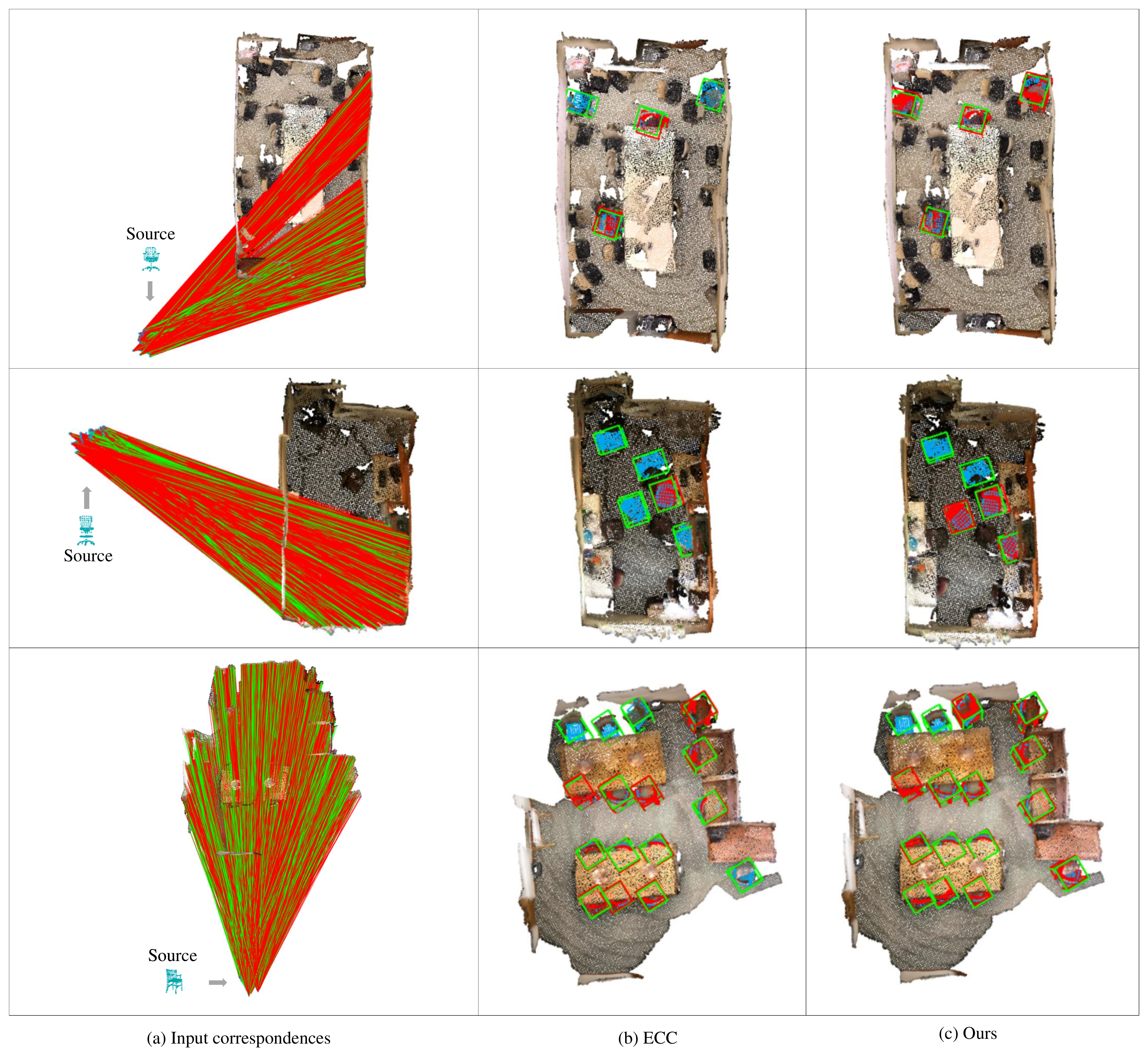}
\caption{\label{fig:12}Visual results on the real dataset.}
\end{figure*}

More visualizations on the synthetic and real datasets are shown in Fig. \ref{fig:11} and Fig. \ref{fig:12}, respectively. We compare with ECC and find that our IBI-S2DC successfully registers more instances in the presence of heavy outliers.

%% file: main.bbl
\begin{thebibliography}{41}
\providecommand{\natexlab}[1]{#1}
\providecommand{\url}[1]{\texttt{#1}}
\expandafter\ifx\csname urlstyle\endcsname\relax
  \providecommand{\doi}[1]{doi: #1}\else
  \providecommand{\doi}{doi: \begingroup \urlstyle{rm}\Url}\fi

\bibitem[Ao et~al.(2021)Ao, Hu, Yang, Markham, and Guo]{SpinNet}
Sheng Ao, Qingyong Hu, Bo Yang, Andrew Markham, and Yulan Guo.
\newblock Spinnet: Learning a general surface descriptor for 3d point cloud registration.
\newblock In \emph{Proc. of the IEEE/CVF Conference on Computer Vision and Pattern Recognition}, pages 11753--11762, 2021.

\bibitem[Avetisyan et~al.(2019)Avetisyan, Dahnert, Dai, Savva, Chang, and Nießner]{Scan2CAD}
Armen Avetisyan, Manuel Dahnert, Angela Dai, Manolis Savva, Angel~X. Chang, and Matthias Nießner.
\newblock Scan2cad: Learning cad model alignment in rgb-d scans.
\newblock In \emph{Proc. of the IEEE/CVF Conference on Computer Vision and Pattern Recognition}, pages 2609--2618, 2019.

\bibitem[Bai et~al.(2021)Bai, Luo, Zhou, Chen, Li, Hu, Fu, and Tai]{PointDSC}
Xuyang Bai, Zixin Luo, Lei Zhou, Hongkai Chen, Lei Li, Zeyu Hu, Hongbo Fu, and Chiew-Lan Tai.
\newblock Pointdsc: Robust point cloud registration using deep spatial consistency.
\newblock In \emph{Proc. of the IEEE/CVF Conference on Computer Vision and Pattern Recognition}, pages 15859--15869, 2021.

\bibitem[Barath and Matas(2018)]{Graph-cut}
Daniel Barath and Ji{\v{r}}{\'\i} Matas.
\newblock Graph-cut ransac.
\newblock In \emph{Proc. of the IEEE Conference on Computer Vision and Pattern Recognition}, pages 6733--6741, 2018.

\bibitem[Barath and Matas(2019)]{Progressive-X}
Daniel Barath and Jiri Matas.
\newblock Progressive-x: Efficient, anytime, multi-model fitting algorithm.
\newblock In \emph{Proc. of the IEEE/CVF International Conference on Computer Vision}, pages 3780--3788, 2019.

\bibitem[Bustos and Chin(2017)]{GORE}
Alvaro~Parra Bustos and Tat-Jun Chin.
\newblock Guaranteed outlier removal for point cloud registration with correspondences.
\newblock \emph{IEEE Transactions on Pattern Analysis and Machine Intelligence}, 40\penalty0 (12):\penalty0 2868--2882, 2017.

\bibitem[Chen and Bhanu(2007)]{GC}
Hui Chen and Bir Bhanu.
\newblock 3d free-form object recognition in range images using local surface patches.
\newblock \emph{Pattern Recognition Letters}, 28\penalty0 (10):\penalty0 1252--1262, 2007.

\bibitem[Chen et~al.(2022)Chen, Sun, Yang, and Tao]{SC2}
Zhi Chen, Kun Sun, Fan Yang, and Wenbing Tao.
\newblock Sc2-pcr: A second order spatial compatibility for efficient and robust point cloud registration.
\newblock In \emph{Proc. of the IEEE/CVF Conference on Computer Vision and Pattern Recognition}, pages 13221--13231, 2022.

\bibitem[Choy et~al.(2020)Choy, Dong, and Koltun]{DGR}
Christopher Choy, Wei Dong, and Vladlen Koltun.
\newblock Deep global registration.
\newblock In \emph{Proc. of the IEEE/CVF Conference on Computer Vision and Pattern Recognition}, pages 2514--2523, 2020.

\bibitem[Dai et~al.(2017)Dai, Chang, Savva, Halber, Funkhouser, and Nießner]{Scannet}
Angela Dai, Angel~X. Chang, Manolis Savva, Maciej Halber, Thomas Funkhouser, and Matthias Nießner.
\newblock Scannet: Richly-annotated 3d reconstructions of indoor scenes.
\newblock In \emph{Proc. of the IEEE Conference on Computer Vision and Pattern Recognition}, pages 2432--2443, 2017.

\bibitem[Drost et~al.(2010)Drost, Ulrich, Navab, and Ilic]{PPF}
Bertram Drost, Markus Ulrich, Nassir Navab, and Slobodan Ilic.
\newblock Model globally, match locally: Efficient and robust 3d object recognition.
\newblock In \emph{IEEE Computer Society Conference on Computer Vision and Pattern Recognition}, pages 998--1005. IEEE, 2010.

\bibitem[Fischler and Bolles(1981)]{RANSAC}
Martin~A Fischler and Robert~C Bolles.
\newblock Random sample consensus: a paradigm for model fitting with applications to image analysis and automated cartography.
\newblock \emph{Communications of the ACM}, 24\penalty0 (6):\penalty0 381--395, 1981.

\bibitem[Glent~Buch et~al.(2014)Glent~Buch, Yang, Kruger, and Gordon~Petersen]{SI}
Anders Glent~Buch, Yang Yang, Norbert Kruger, and Henrik Gordon~Petersen.
\newblock In search of inliers: 3d correspondence by local and global voting.
\newblock In \emph{Proc. of the IEEE Conference on Computer Vision and Pattern Recognition}, pages 2067--2074, 2014.

\bibitem[Guo et~al.(2021)Guo, Xing, Quan, Yan, Gu, Liu, and Zhang]{PPFVoting}
Jianwei Guo, Xuejun Xing, Weize Quan, Dong-Ming Yan, Qingyi Gu, Yang Liu, and Xiaopeng Zhang.
\newblock Efficient center voting for object detection and 6d pose estimation in 3d point cloud.
\newblock \emph{IEEE Transactions on Image Processing}, 30:\penalty0 5072--5084, 2021.

\bibitem[Huang et~al.(2021)Huang, Gojcic, Usvyatsov, Wieser, and Schindler]{Predator}
Shengyu Huang, Zan Gojcic, Mikhail Usvyatsov, Andreas Wieser, and Konrad Schindler.
\newblock Predator: Registration of 3d point clouds with low overlap.
\newblock In \emph{Proc. of the IEEE/CVF Conference on Computer Vision and Pattern Recognition}, pages 4267--4276, 2021.

\bibitem[Kluger et~al.(2020)Kluger, Brachmann, Ackermann, Rother, Yang, and Rosenhahn]{CONSAC}
Florian Kluger, Eric Brachmann, Hanno Ackermann, Carsten Rother, Michael~Ying Yang, and Bodo Rosenhahn.
\newblock Consac: Robust multi-model fitting by conditional sample consensus.
\newblock In \emph{Proc. of the IEEE/CVF Conference on Computer Vision and Pattern Recognition}, pages 4634--4643, 2020.

\bibitem[Lee et~al.(2021)Lee, Kim, Cho, and Park]{DHVG}
Junha Lee, Seungwook Kim, Minsu Cho, and Jaesik Park.
\newblock Deep hough voting for robust global registration.
\newblock In \emph{Proc. of the IEEE/CVF International Conference on Computer Vision}, pages 15994--16003, 2021.

\bibitem[Leordeanu and Hebert(2005)]{ST}
Marius Leordeanu and Martial Hebert.
\newblock A spectral technique for correspondence problems using pairwise constraints.
\newblock In \emph{Proc. International Conference on Computer Vision}, pages 1482--1489. IEEE, 2005.

\bibitem[Lowe(2004)]{NNSR}
David~G Lowe.
\newblock Distinctive image features from scale-invariant keypoints.
\newblock \emph{International Journal of Computer Vision}, 60:\penalty0 91--110, 2004.

\bibitem[Magri and Fusiello(2014)]{T-Linkage}
Luca Magri and Andrea Fusiello.
\newblock T-linkage: A continuous relaxation of j-linkage for multi-model fitting.
\newblock In \emph{Proc. of the IEEE Conference on Computer Vision and Pattern Recognition}, pages 3954--3961, 2014.

\bibitem[Magri and Fusiello(2016)]{RansaCov}
Luca Magri and Andrea Fusiello.
\newblock Multiple model fitting as a set coverage problem.
\newblock In \emph{Proc. of the IEEE Conference on Computer Vision and Pattern Recognition}, pages 3318--3326, 2016.

\bibitem[Otsu(1979)]{OSTU}
Nobuyuki Otsu.
\newblock A threshold selection method from gray-level histograms.
\newblock \emph{IEEE Transactions on Systems, Man, and Cybernetics}, 9\penalty0 (1):\penalty0 62--66, 1979.

\bibitem[Parra et~al.(2019)Parra, Chin, Neumann, Friedrich, and Katzmann]{PMC}
Alvaro Parra, Tat-Jun Chin, Frank Neumann, Tobias Friedrich, and Maximilian Katzmann.
\newblock A practical maximum clique algorithm for matching with pairwise constraints.
\newblock \emph{arXiv preprint arXiv:1902.01534}, 2019.

\bibitem[Qi et~al.(2017)Qi, Yi, Su, and Guibas]{Pointnet++}
Charles~Ruizhongtai Qi, Li Yi, Hao Su, and Leonidas~J Guibas.
\newblock Pointnet++: Deep hierarchical feature learning on point sets in a metric space.
\newblock \emph{Advances in Neural Information Processing Systems}, 30, 2017.

\bibitem[Quan and Yang(2020)]{CG}
Siwen Quan and Jiaqi Yang.
\newblock Compatibility-guided sampling consensus for 3-d point cloud registration.
\newblock \emph{IEEE Transactions on Geoscience and Remote Sensing}, 58\penalty0 (10):\penalty0 7380--7392, 2020.

\bibitem[Quan et~al.(2018)Quan, Ma, Hu, Fang, and Ma]{OP}
Siwen Quan, Jie Ma, Fangyu Hu, Bin Fang, and Tao Ma.
\newblock Local voxelized structure for 3d binary feature representation and robust registration of point clouds from low-cost sensors.
\newblock \emph{Information Sciences}, 444:\penalty0 153--171, 2018.

\bibitem[Rodol{\`a} et~al.(2013)Rodol{\`a}, Albarelli, Bergamasco, and Torsello]{GTM}
Emanuele Rodol{\`a}, Andrea Albarelli, Filippo Bergamasco, and Andrea Torsello.
\newblock A scale independent selection process for 3d object recognition in cluttered scenes.
\newblock \emph{International Journal of Computer Vision}, 102:\penalty0 129--145, 2013.

\bibitem[Savva et~al.(2016)Savva, Yu, Su, Aono, Chen, Cohen-Or, Deng, Su, Bai, Bai, et~al.]{ShapeNet}
Manolis Savva, Fisher Yu, Hao Su, M Aono, B Chen, D Cohen-Or, W Deng, Hang Su, Song Bai, Xiang Bai, et~al.
\newblock Shrec16 track: largescale 3d shape retrieval from shapenet core55.
\newblock In \emph{Proc. of the Eurographics Workshop on 3D Object Retrieval}, 2016.

\bibitem[Tang and Zou(2022)]{ECC}
Weixuan Tang and Danping Zou.
\newblock Multi-instance point cloud registration by efficient correspondence clustering.
\newblock In \emph{Proc. of the IEEE/CVF Conference on Computer Vision and Pattern Recognition}, pages 6667--6676, 2022.

\bibitem[Tombari and Di~Stefano(2010)]{3DHV}
Federico Tombari and Luigi Di~Stefano.
\newblock Object recognition in 3d scenes with occlusions and clutter by hough voting.
\newblock In \emph{Pacific-Rim Symposium on Image and Video Technology}, pages 349--355. IEEE, 2010.

\bibitem[Weibull(1997)]{Evolutionary}
J{\"o}rgen~W Weibull.
\newblock \emph{Evolutionary game theory}.
\newblock MIT press, 1997.

\bibitem[Yang et~al.(2020)Yang, Shi, and Carlone]{TEASER}
Heng Yang, Jingnan Shi, and Luca Carlone.
\newblock Teaser: Fast and certifiable point cloud registration.
\newblock \emph{IEEE Transactions on Robotics}, 37\penalty0 (2):\penalty0 314--333, 2020.

\bibitem[Yang et~al.(2019)Yang, Xiao, Cao, and Yang]{CV}
Jiaqi Yang, Yang Xiao, Zhiguo Cao, and Weidong Yang.
\newblock Ranking 3d feature correspondences via consistency voting.
\newblock \emph{Pattern Recognition Letters}, 117:\penalty0 1--8, 2019.

\bibitem[Yang et~al.(2021{\natexlab{a}})Yang, Huang, Quan, Qi, and Zhang]{SACCOT}
Jiaqi Yang, Zhiqiang Huang, Siwen Quan, Zhaoshuai Qi, and Yanning Zhang.
\newblock Sac-cot: Sample consensus by sampling compatibility triangles in graphs for 3-d point cloud registration.
\newblock \emph{IEEE Transactions on Geoscience and Remote Sensing}, 60:\penalty0 1--15, 2021{\natexlab{a}}.

\bibitem[Yang et~al.(2021{\natexlab{b}})Yang, Huang, Quan, Zhang, Zhang, and Cao]{Metrics}
Jiaqi Yang, Zhiqiang Huang, Siwen Quan, Qian Zhang, Yanning Zhang, and Zhiguo Cao.
\newblock Toward efficient and robust metrics for ransac hypotheses and 3d rigid registration.
\newblock \emph{IEEE Transactions on Circuits and Systems for Video Technology}, 32\penalty0 (2):\penalty0 893--906, 2021{\natexlab{b}}.

\bibitem[Yang et~al.(2021{\natexlab{c}})Yang, Xian, Wang, and Zhang]{CorresGrouping}
Jiaqi Yang, Ke Xian, Peng Wang, and Yanning Zhang.
\newblock A performance evaluation of correspondence grouping methods for 3d rigid data matching.
\newblock \emph{IEEE Transactions on Pattern Analysis and Machine Intelligence}, 43\penalty0 (6):\penalty0 1859--1874, 2021{\natexlab{c}}.

\bibitem[Yang et~al.(2022)Yang, Chen, Quan, Wang, and Zhang]{LnT}
Jiaqi Yang, Jiahao Chen, Siwen Quan, Wei Wang, and Yanning Zhang.
\newblock Correspondence selection with loose--tight geometric voting for 3-d point cloud registration.
\newblock \emph{IEEE Transactions on Geoscience and Remote Sensing}, 60:\penalty0 1--14, 2022.

\bibitem[Yuan et~al.(2022)Yuan, Li, Jin, Chen, and Wang]{PointCLM}
Mingzhi Yuan, Zhihao Li, Qiuye Jin, Xinrong Chen, and Manning Wang.
\newblock Pointclm: A contrastive learning-based framework for multi-instance point cloud registration.
\newblock In \emph{Proc. of the European Conference on Computer Vision}, pages 595--611. Springer, 2022.

\bibitem[Zeng et~al.(2017)Zeng, Song, Nie{\ss}ner, Fisher, Xiao, and Funkhouser]{3DMatch}
Andy Zeng, Shuran Song, Matthias Nie{\ss}ner, Matthew Fisher, Jianxiong Xiao, and Thomas Funkhouser.
\newblock 3dmatch: Learning local geometric descriptors from rgb-d reconstructions.
\newblock In \emph{Proc. of the IEEE Conference on Computer Vision and Pattern Recognition}, pages 1802--1811, 2017.

\bibitem[Zhang et~al.(2023)Zhang, Yang, Zhang, and Zhang]{MAC}
Xiyu Zhang, Jiaqi Yang, Shikun Zhang, and Yanning Zhang.
\newblock 3d registration with maximal cliques.
\newblock In \emph{Proc. of the IEEE/CVF Conference on Computer Vision and Pattern Recognition}, pages 17745--17754, 2023.

\bibitem[Zhou et~al.(2016)Zhou, Park, and Koltun]{FGR}
Qian-Yi Zhou, Jaesik Park, and Vladlen Koltun.
\newblock Fast global registration.
\newblock In \emph{Proc. of the European Conference on Computer Vision}, pages 766--782. Springer, 2016.

\end{thebibliography}
